\documentclass[11pt]{article}

\usepackage[final]{acl}

\usepackage{times}
\usepackage{latexsym}

\usepackage[T1]{fontenc}
\usepackage[utf8]{inputenc}

\usepackage{microtype}

\usepackage{inconsolata}

\usepackage{graphicx}
\usepackage{booktabs}
\usepackage{soul}
\usepackage{xcolor}
\usepackage{makecell}
\usepackage{adjustbox}
\usepackage{float}
\usepackage{multirow} 
 \usepackage{pgfplots}
\usepackage[table]{xcolor}
\usepackage{arydshln}
\usepackage{amsmath}
\usepackage{longtable}
\usepackage{listings}
\usepackage{subcaption}
\usepackage[table]{xcolor}
\usepackage{amsfonts}  % for \mathbb
\usepackage{enumitem}

\title{Linear Probing Provides Robust and Efficient Detection of Machine-Generated Text}

\author{%
  Gerrit Quaremba$^{1}$, Hanqi Yan$^{1}$, Elizabeth Black$^{1}$, Denny Vrandečić$^{2}$, Elena Simperl$^{1}$ \\
  $^{1}$King's College London, 
  $^{2}$Wikimedia Foundation \\
  \texttt{\{gerrit.quaremba,hanqi.yan,elizabeth.black,elena.simperl\}@kcl.ac.uk} \\\texttt{denny@wikimedia.org}
}

\begin{document}
\maketitle

% Write in Neal Nanda style; Anthropic sleeper agents also has a very nice intro
% \begin{itemize}
%     \item ad takeaway to tables/figures
%     \item Story in terms of \textit{robustness} and \textit{efficiency} --- mentio these two in the abstract with numbers
%     \item learn to distance has great tables and structure
%     \item Great intro plot: https://openreview.net/pdf?id=C5Jj3QKQav + we should add the frozen llm and that we retrieve activations
%     \item \hl{Hanqi}: when we pca activations, we do get bascially them same performance with k=100 then with full dimensions (4,096). Is there any reason to do it/not to do it? I initially wanted to make the argument, that MGT lives in a low-dim subspace. But after I saw that applying PCA does not improve AUC, I dropped it. I believe that this is because the L1 penalty effectively shrink many dimensions to zero. 

%     \item Wer hsopt that rank thing. Based on these insights, we shot that MGT probes, simple linear probes that detect MGT, are ...

\begin{abstract}
Distinguishing machine-generated text (MGT) from human-written text (HWT) becomes increasingly important due to potential misuse.
However, most supervised detectors often degrade out-of-domain (OOD) and require large, diverse training sets.
In this work, we analyze the linearity and quality of MGT representations and show that simple linear probes outperform a wide range of detectors while being substantially more sample-efficient.
We first show that MGT and HWT latent representations are linearly separable in low-dimensional space, and provide a plausible explanation for this separability through systematic differences in their representation quality.
Motivated by these insights, we train two variants of simple linear probes and evaluate them across 4 benchmarks against 16 baselines.
Probes consistently improve OOD detection (+11 AUC), requiring solely ${<}100$ samples to reach near-peak performance.
We show that this transferability arises because probes recover a shared latent MGT direction that generalizes across diverse settings.
Finally, we demonstrate that probing vectors capture a continuous spectrum of ``machineness'', highlighting their potential for fine-grained estimation of AI-edited text.
Overall, our work provides insights into latent-space differences between MGT and HWT and demonstrates the potential of linear probes as as robust and sample-efficient MGT detectors.
We release our code on~\href{https://github.com/gerritq/mgt_probes}{github}.
\end{abstract}

%%%%%%%%%%%%%%%%%%%%%%%%%%%%%%%%%%%%%%%%%%%%%%%%%%%%%%%%%%%%%%%%%%%%%%%%
% INTRODUCTION
%%%%%%%%%%%%%%%%%%%%%%%%%%%%%%%%%%%%%%%%%%%%%%%%%%%%%%%%%%%%%%%%%%%%%%%%

\section{Introduction}

Large language models (LLMs) now generate human-like text at scale, and their output has spread rapidly across platforms such as social media~\cite{sun2025we} or academic writing~\cite{liang2024monitor}.
This proliferation raises the risk of potential misuse, including misinformation campaigns~\cite{goldstein2023generative} or plagiarism~\cite{hutson2024rethinking}.
In response, a growing line of work builds detectors that separate machine-generated text (MGT) from human-written text (HWT)~\cite{crothers2023machine,wu2025survey}.

Prior work groups MGT detectors into two families~\cite{wu2025survey}: zero-shot methods~\cite[e.g.,][]{hans2024binoculars,bao2023fast} and supervised classifiers~\cite[e.g.,][]{hu2023radar,verma2024ghostbuster}.\footnote{We focus on passive detection and exclude watermarking.}
While supervised approaches dominate recent benchmarks~\cite{macko2025multi,quaremba2026tsm}, their practical utility remains limited: they often generalize poorly to out-of-domain (OOD) settings~\cite{wu2025survey,doughman2025exploring} and require large, diverse training corpora~\cite{hu2023radar}.
Additionally, a recent line of work leverages how LLMs rewrite machine- versus human-written text differently~\cite{mao2024raidar,wu2025wrote}, but adds substantial overhead from generating rewrites.

A promising direction for overcoming these limitations lies in the Linear Representation Hypothesis (LRH)~\cite{park2023linear}, which posits that high-level concepts are encoded as linear directions in latent space.
Recent work has identified such directions for concepts including truth~\cite{marks2024the} or political ideology~\cite{kim2025linear}.
This raises a natural question: \textit{Are machine- and human-written text linearly separable in latent space, and if so, why? How can we exploit this structure for enhanced MGT detection?}

In this work, we empirically study the latent geometry of HWT and MGT and show that simple linear probes are sufficient to improve the robustness and sample efficiency of MGT detection.
We begin by visualizing that HWT and MGT are largely linearly separable in low-dimensional space, which we corroborate by \textit{reduced} detection performance when increasing the complexity of nonlinear probes (\S\ref{sec:mgt_rep:linear}).
We then investigate \textit{why} this linearity emerges by analyzing both populations with four representation quality metrics~\cite{skean2025layer}:
compared to HWT, MGT representations collapse into a compressed, more anisotropic, and lower-dimensional subspace (\S\ref{sec:mgt_rep:qual}).
These systematic differences provide a plausible explanation for why a simple hyperplane can effectively separate the two in latent space.

Leveraging these insights, we train \textbf{MGT probes}: simple linear probes~\cite{alain2017understanding,nanda2023emergent} on frozen LM representations to detect MGT.
We consider two variants: layer-wise probes and a single probe trained on representations concatenated across layers. 
Crucially, both variants operate on PCA-reduced features, retaining only 2.4\% of the original dimensions with minimal performance loss.
The resulting probing vector defines a linear ``machineness'' direction in latent space, which we use for detection by projecting test activations onto it~\cite{hollinsworth2024language}.
Across 4 benchmarks spanning 16 evaluation settings and compared against 16 baseline detectors, MGT probes improve in-domain detection by up to 18 AUC (\S\ref{sec:exp:id}) and 11 AUC OOD (\S\ref{sec:exp:ood}).

To explain this strong generalization, we analyze probing vectors learned across diverse datasets and settings. 
We find that the vectors are highly aligned, indicating that probes recover a shared latent MGT direction (\S\ref{sec:exp:probes}). 
Moreover, probes achieve strong performance with only 10--100 training samples and exhibit substantially lower sampling uncertainty than training-based baselines (\S\ref{sec:exp:sample}). 
Together, our findings suggest that MGT is encoded as a compact, shared, and linearly accessible direction in latent space.
This allows simple linear probes to generalize better while requiring fewer examples than more complex supervised detectors.

Lastly, we examine whether probing vectors can serve as a continuous measure of AI editing in text~\cite{zhang2024llm} (\S\ref{sec:exp:edit}). We find strong correlations between probe projection scores and edit-strength metrics~\cite{saha2025apt}, meaning that more extensively AI-edited text is projected closer to the MGT subspace. This finding suggests that the latent MGT direction encodes a \textit{continuous} spectrum of ``machineness,'' which can be leveraged for fine-grained estimation of AI-edited text.

Our contributions are threefold:

\begin{enumerate}[itemsep=2pt, topsep=2pt, parsep=0pt]
    \item We find that human- and machine-written text are linearly separable in latent space and explain this separability through systematic differences in their representation geometry.

    \item We introduce two simple linear probe variants that recover this latent direction, improving OOD generalization over supervised detectors (+11 AUC) while exhibiting greater reliability and requiring only 10--100 training examples.

    \item We show that the learned MGT direction encodes a \textit{continuous} spectrum of ``machineness,'' enabling fine-grained estimation of AI editing strength through probe projection scores.
\end{enumerate}

\begin{figure*}[h]
     \centering
     \includegraphics[width=1\linewidth]{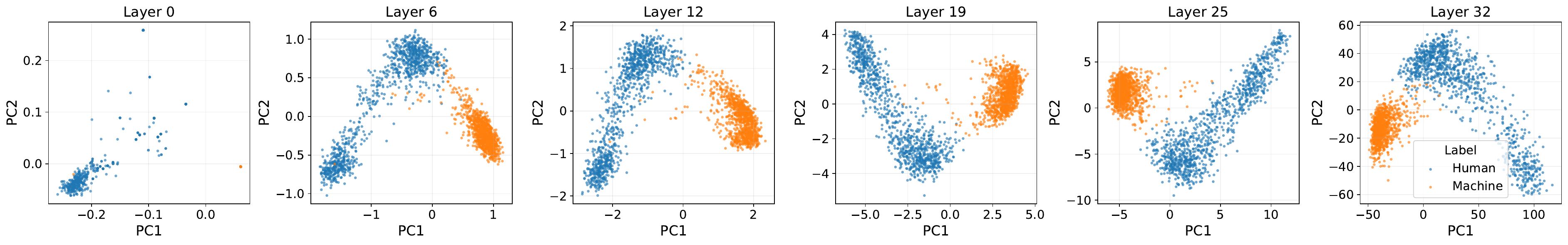}
    \caption{Projections of Wikipedia human- and machine-text hidden states onto the first two principal components across layers. \textbf{Takeaway}: Machine- and human-text representations become linearly separable as early as layer 6, and this separation remains consistent through the final layers. Appendix~\ref{app:mgt_rep} confirms the same pattern across other domains.}
    \label{fig:layer_pca}
\end{figure*}

\section{Related Work}

\paragraph{Linear Representation Hypothesis (LRH)}
Recent work has provided growing support for the LRH, which posits that high-level concepts are represented as linear directions in latent space~\cite{park2023linear}.
Various studies have successfully used linear probing to identify concepts such as spatial reasoning~\cite{nanda2023emergent}, truth~\cite{marks2023geometry}, sentiment~\cite{hollinsworth2024language}, or political ideology~\cite{kim2025linear}.
To identify probing vectors, one typically construct a contrastive dataset containing positive and negative instances of the concept of interest (e.g., MGT and HWT).
Common approaches for identifying probing directions include difference-in-means~\cite{belrose2023diffinmeans} and linear probes~\cite{alain2017understanding}.
The resulting probing vector represents the concept as a linear direction in activation space and can be used for detection, interpretation, or representation steering~\cite{zou2025down}.

    % \item Show that other works like Goodifre find non-linear concetps 
    % (https://www.goodfire.ai/research/manifold-steering)

\paragraph{MGT Detectors}
MGT detectors can be broadly categorized as zero-shot or supervised methods~\cite{wu2024detectrl}.
Zero-shot detectors are training-free and typically exploit discriminative differences in token probabilities~\cite{bao2023fast,su2023detectllm,hans2024binoculars} or leverage the observation that LLMs rewrite MGT and HWT differently~\cite{wu2025wrote,zhu2023beat,yang2024dna}.
Most supervised detectors formulate MGT detection as a binary classification task and train machine-learning classifiers~\cite{hu2023radar,guo2024biscope,mao2024raidar,thai2026editlens}.
Compared to prior latent-representation-based detectors, our approach operates directly on activations rather than a single intrinsic-dimensionality feature~\cite{tulchinskii2023intrinsic}, uses \textit{linear} probes across layers rather than a single-layer \textit{nonlinear} classifier~\cite{yu2024fluo}, and learns probes in a \textit{low-dimensional} space rather than identifying directions in a \textit{training-free} manner in the \textit{full} representation space~\cite{chen2025repreguard}.
We review current MGT detectors and provide a detailed comparison with related work in Appendix~\ref{app:exp_setup}.

%%%%%%%%%%%%%%%%%%%%%%%%%%%%%%%%%%%%%%%%%%%%%%%%%%%%%%%%%%%%%%%%%%%%%%%%
% Why do representations differ
%%%%%%%%%%%%%%%%%%%%%%%%%%%%%%%%%%%%%%%%%%%%%%%%%%%%%%%%%%%%%%%%%%%%%%%%

\begin{figure}[t]
     \centering
     \begin{subfigure}[b]{.8\linewidth}
         \centering
         \includegraphics[width=\linewidth]{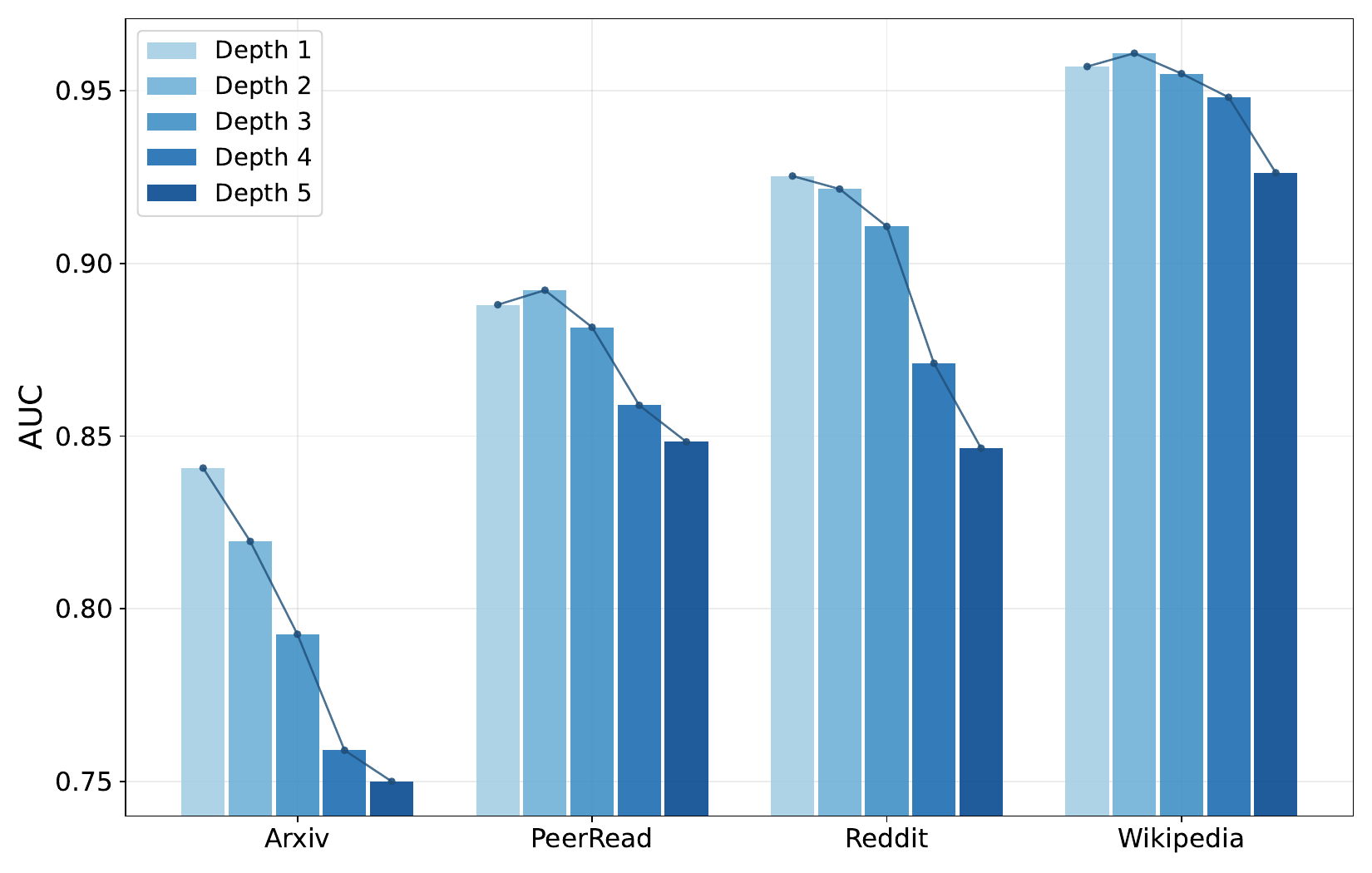}
     \end{subfigure}
     
     \vspace{1em} % Add some vertical spacing between them

     \begin{subfigure}[b]{.8\linewidth}
         \centering
         \includegraphics[width=\linewidth]{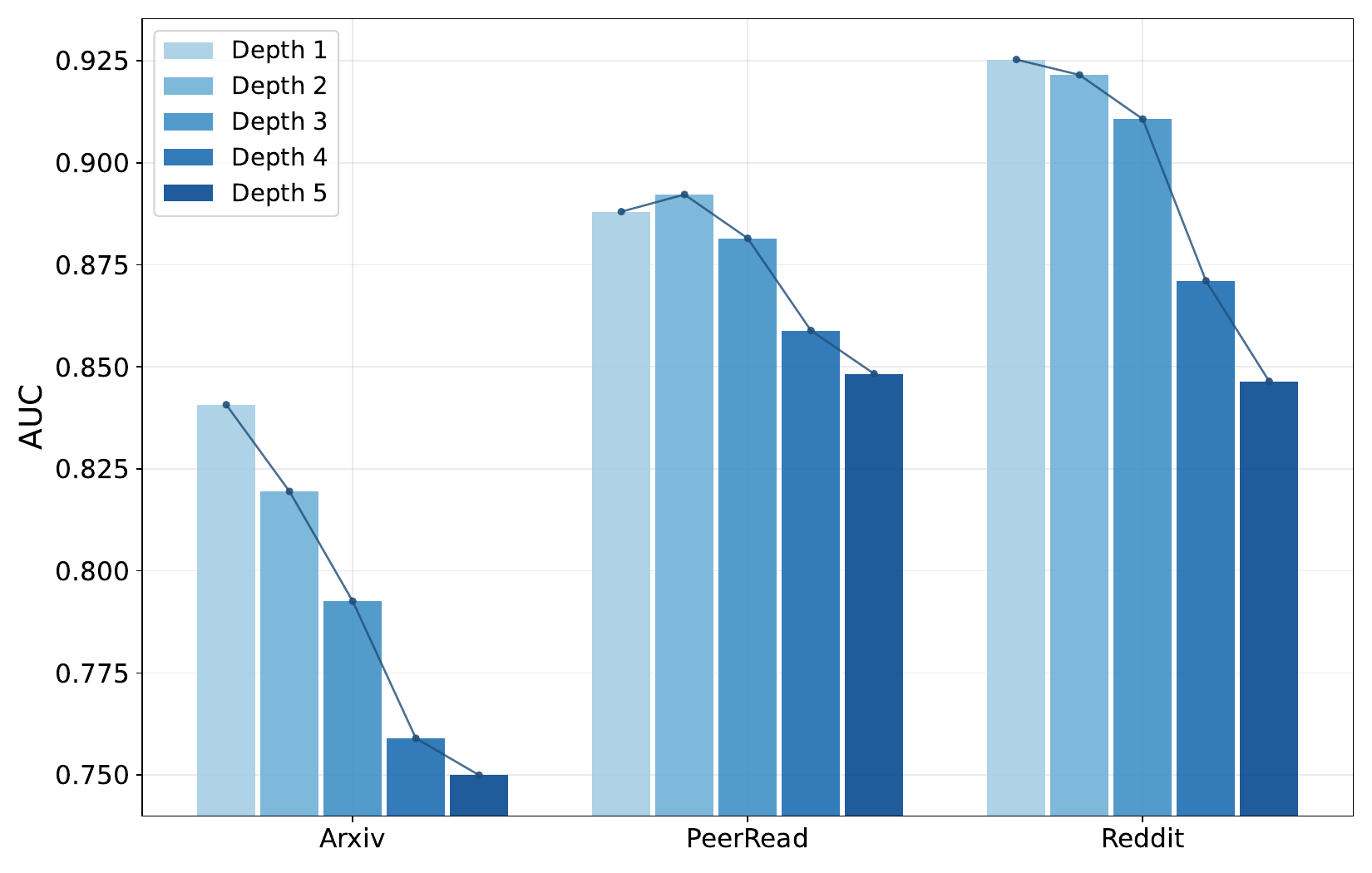}
     \end{subfigure}
     
    \caption{In-domain (top) and OOD (bottom) AUC scores for probing middle-layer representations with MLPs of increasing levels of non-linearity. \textbf{Takeaway}: Increasing model complexity consistently reduces performance, supporting the hypothesis that machine- and human-text representations are \textit{linearly} separable in latent space.}
     \label{fig:map}
\end{figure}

\section{On the Separability and Quality of MGT Representations}
\label{sec:mgt_rep}

In this section, we first examine the \textit{linear separability} of HWT and MGT representations (\S\ref{sec:mgt_rep:linear}) and then analyze their \textit{representation quality} to characterize their latent-space differences (\S\ref{sec:mgt_rep:qual}). 
We conduct all analyses on the Wikipedia subset of M4GT~\cite{wang2024m4gt} using \texttt{Llama-8B}~\cite{grattafiori2024llama3} as the backbone model. Appendix~\ref{app:mgt_rep:results} shows that our findings generalize across domains.

\begin{figure*}[h]
     \centering
     \includegraphics[width=1\linewidth]{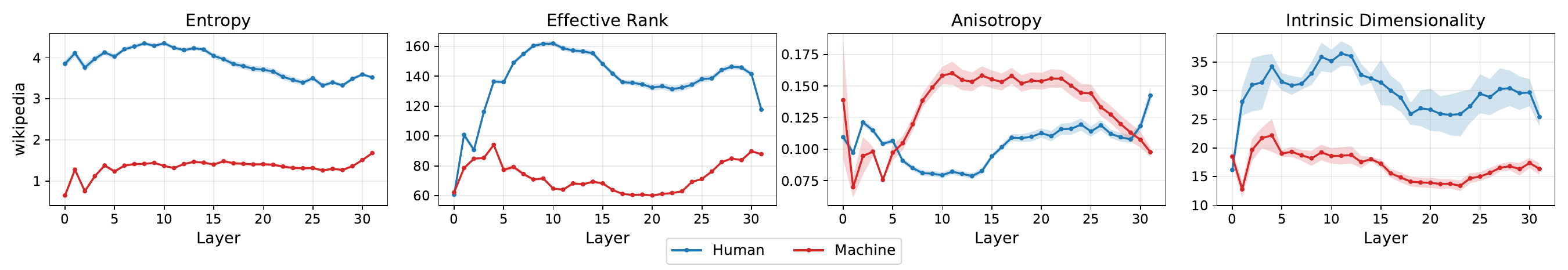}
    \caption{Representation-quality metrics for human and machine latent representations from the perspectives of information content (Entropy, Effective Rank) and geometric structure (Anisotropy, Intrinsic Dimensionality). \textbf{Takeaway}: Machine-text representations are more compressed, concentrated along fewer dominant directions, and located on simpler manifolds than human-written text representations. These structural differences help explain why the two are linearly separable in latent space.}
    \label{fig:f_qual_wiki}
\end{figure*}

\subsection{Linear Separability}
\label{sec:mgt_rep:linear}

We begin by asking: \textit{Are HWT and MGT linearly separable in latent space?}
While prior work has shown that linear classifiers can work~\cite{tulchinskii2023intrinsic,chen2025repreguard},\footnote{We have included the detailed comparison with them in Appendix~\ref{app:exp_setup:comp}.} it remains unclear whether the underlying decision boundary is truly linear or whether more complex nonlinear boundaries are required.

\textbf{MGT and HWT are linearly separable in low-dimensional latent spaces.}
Figure~\ref{fig:layer_pca} visualizes residual stream activations of MGT and HWT across layers, projected onto the first two principal components.
While the two distributions overlap in the first layer, they become clearly \textit{linearly} separable as early as layer 6.
This separation remains remarkably stable throughout the remaining layers.
% Appendix Figure~\ref{app:fig:pca_domains} further shows that this linear structure persists across domains, and 
Appendix Figure~\ref{app:fig:map} further shows that each domain and generator forms distinct local subspaces, within which HWT and MGT remain linearly separable.

To test whether the decision boundary is indeed linear, we train MLP probes of increasing complexity on mid-layer representations to predict MGT~\cite{gurnee2024language}.
If nonlinear structure were important for detection, deeper classifiers should improve performance.
Instead, Figure~\ref{fig:map} shows that detection performance deteriorates as nonlinearity increases in both ID and OOD settings.
Our findings empirically support that MGT and HWT are separated by a predominantly linear decision boundary in latent space.

\subsection{Representation Quality}
\label{sec:mgt_rep:qual}

Yet, it remains unclear which representational differences make this linear separation possible. We therefore ask: \textit{How do HWT and MGT differ in latent space?}
To this end, we analyze their latent representations from the perspective of \textit{information content} (Entropy, Effective Rank) and \textit{geometric structure} (Anisotropy, Intrinsic Dimension) representation quality metrics~\cite{skean2025layer}:
(\textbf{1})~\textbf{Entropy}~\cite{giraldo2015entropy} measures how much information a representation contains; 
(\textbf{2})~\textbf{Effective Rank}~\cite{roy2007effective} quantifies the amount of non-redundant information in a representations;
(\textbf{3})~\textbf{Anisotropy}~\cite{razzhigaev2024shape} measures the degree to which activations concentrate along specific directions in latent space; and 
(\textbf{4})~\textbf{Intrinsic Dimension}~\cite{Facco2017estimating} estimates the minimum number of dimensions required to describe the geometric structure of representations.
Appendix~\ref{app:mgt_rep:metrics} provides additional intuition and formulae.

\textbf{MGT representations are more compressed, while HWT representations contain richer and more diverse features.}
Figure~\ref{fig:f_qual_wiki} plots representation-quality metrics across layers for HWT and MGT activations.
The information-theoretic metrics (Entropy and Effective Rank) capture how broadly information is distributed across representations.
Higher values indicate that variance is spread across many dimensions, whereas lower values indicate that information collapses into fewer dominant components.
Across layers, HWT exhibits substantially higher entropy than MGT, suggesting that human-written text activates a broader and less redundant set of features.
In contrast, MGT has much lower entropy, indicating a more compressed representation in which activations carry less diverse information.
With the same pattern appearing for Effective Rank, both metrics suggest that MGT representations occupy a more information-compressed subspace, while HWT retains richer, more heterogeneous latent features.

\textbf{MGT occupies a narrower and lower-dimensional activation geometry than HWT.}
The geometric metrics (Anisotropy and Intrinsic Dimension) characterize the shape of activation distributions.
Higher anisotropy indicates that representations align strongly with a small number of dominant directions, whereas higher intrinsic dimensionality indicates that more dimensions are required to capture their local structure.
Across most middle and later layers, MGT activations are more anisotropic than HWT activations, suggesting that MGT concentrates along a narrower set of directions in latent space.
At the same time, MGT exhibits consistently lower intrinsic dimensionality, indicating that its representations lie on a simpler, lower-dimensional manifold.
HWT shows the opposite pattern: lower anisotropy and higher intrinsic dimensionality suggest that its activations are more uniformly distributed and occupy a more complex manifold.
This finding is consistent with prior work reporting lower intrinsic dimensionality for MGT~\cite{tulchinskii2023intrinsic}. We further show that this structure emerges systematically across layers and remains consistent across diverse domains.

% Differences across all four metrics are most pronounced in the middle layers.
% This suggests that MGT-specific structure emerges once representations become sufficiently abstract, but before the final layers become more specialized toward next-token prediction.
\textbf{Takeaway}
Our linearity and representation analyses support three takeaways:
(\textit{1}) HWT and MGT representations are linearly separable in low-dimensional latent space beginning early layers;
(\textit{2}) information-content metrics show that MGT representations are more compressed, whereas HWT representations are richer and more diverse; and
(\textit{3}) geometric metrics show that MGT representations occupy a narrower, more anisotropic, and lower-dimensional manifold, whereas HWT spans a broader and more complex region of activation space.
These structural differences provide a plausible explanation of \textit{why} simple linear MGT detectors likely are sufficient and effective in latent space.

\section{Methodology}
\label{sec:method}

The linear separation between MGT and HWT representations, driven by systematic distributional differences in latent space, motivates the use of \textbf{MGT probes}.
Drawing on the interpretability literature~\cite{alain2017understanding,belinkov2022probing,li2023inference}, these probes are simple linear classifiers trained on frozen hidden-state activations to identify MGT.
We consider two variants: a \textbf{Layer-Averaged Linear Probe (LLP)}, which trains one linear probe per layer and averages predictions~\cite{nordby2026ensemble}; and a \textbf{Concatenated-Layer Linear Probe (CLP)}, which trains a single linear probe on the concatenation of hidden states. 
%We leave the exploration of alternative probes such as attention probes~\cite{mckenzie2026detecting} for future work.

\paragraph{Preliminaries}
Let $x_i = (x_1, \ldots, x_{T_i})$ denote an input sequence of length $T_i$.
An LLM processes $x_i$ through $L$ transformer layers and produces a hidden state $\mathbf{h}_{i,t}^{(\ell)} \in \mathbb{R}^{d}$ for each token position $t \in \{1,\ldots,T_i\}$ and layer $\ell \in \{1,\ldots,L\}$, where $d$ is the hidden-state dimension.
Given a binary MGT dataset, we extract the last-token hidden state from each layer, yielding the probing dataset
$\mathcal{D}_{\mathrm{probe}} = \{(\{\mathbf{h}_{i}^{(\ell)}\}_{\ell=1}^{L}, y_i)\}_{i=1}^{N}$,
where $y_i \in \{0,1\}$ indicates whether an instance is machine-generated ($y_i = 1$).

As Section~\ref{sec:mgt_rep} shows that MGT and HWT representations are linearly separable in low-dimensional spaces, we train our MGT probes on PCA-reduced activations. Specifically, for each layer, we fit PCA on the training activations and project them onto the top 100 principal components. With Llama-3-8B~\cite{grattafiori2024llama3} as the base model, this reduces the representation dimensionality to \(100/4096 \approx 2.4\%\) of the original space.

We ablate this choice in Appendix~\ref{app:ablation}. Probe performance with and without PCA is nearly identical, indicating that PCA does not drive linear separability.

\paragraph{Layer-Averaged Linear Probe (LLP)}
% For the Layer-Averaged Linear Probe, 
We train an independent logistic regression classifier $f(\cdot)$ for each layer $\ell$:
\begin{equation}
    \label{eq:1}
    \min_{\mathbf{w}^{(\ell)}}
    \frac{1}{N}
    \sum_{i=1}^{N}
    \mathcal{L}
    \left(
    y_i,
    f(\mathbf{h}_{i}^{(\ell)}; \mathbf{w}^{(\ell)})
    \right)
    +
    \lambda \|\mathbf{w}^{(\ell)}\|_2^2 .
\end{equation}

where $\lambda$ controls regularization and $\mathcal{L}$ is the binary cross-entropy loss.

After training, we normalize the weight vector $\mathbf{v}^{(\ell)} = \frac{\mathbf{w}^{(\ell)}}{\left\lVert \mathbf{w}^{(\ell)} \right\rVert_2}$, which defines the \textbf{probing vector} for layer $\ell$. 
Intuitively, $\mathbf{v}^{(\ell)}$ captures the linear direction in activation space that separates MGT from HWT~\cite{gurnee2024language}.

At inference time, we pass a test instance $x_{\mathrm{test}}$ through the LLM, extract its last-token hidden state $\mathbf{h}_{\mathrm{test}}^{(\ell)}$ at each layer, and project it onto the corresponding probing direction:
$s_{\mathrm{test}}^{(\ell)}{=}{\mathbf{h}_{\mathrm{test}}^{(\ell)}}^\top \mathbf{v}^{(\ell)}$~\cite{hollinsworth2024language}.
This projection measures how strongly the hidden state points in the MGT direction identified by the probe. We then compute the final LLP score by averaging the layer-wise projection scores:
$s_{\mathrm{LLP}}(x_{\mathrm{test}}) = \frac{1}{L} \sum_{\ell=1}^{L} s_{\mathrm{test}}^{(\ell)}$.
Higher scores indicate stronger alignment with the MGT direction, and vice versa.

% \begin{table*}[h]
%     \centering
%     {\renewcommand{\arraystretch}{1.0}
%     \begin{adjustbox}{width=\linewidth}
%         \input{tables/t_id}
%     \end{adjustbox}}
%     \caption{In-domain detection AUC scores across four benchmarks, comparing 15 baseline detectors (zero-shot and supervised) with MGT probes. LLP = Layer-Averaged Linear Probes. CLP=Concatenated-Layer Probes. For TSM, FP=First Paragraph, PE=Paragraph Extension, SUM=Summarization, and TST=Text Style Transfer. $\Delta$ denotes the gain over the strongest baseline. \textbf{Takeaway}: Both LLP and CLP consistently outperform all baselines by 0.53--18.68 AUC points.}
%     \label{tab:t_id}
% \end{table*}

\begin{table*}[h]
    \centering
    {\renewcommand{\arraystretch}{1.0}
    \begin{adjustbox}{width=1\linewidth}

\begin{tabular}{lcccccccccccccccc}
\toprule
& \multicolumn{4}{c}{\textbf{DetectRL~\citep{wu2024detectrl}}} & \multicolumn{4}{c}{\textbf{MultiSocial~\citep{macko2025multi}}} & \multicolumn{4}{c}{\textbf{RAID~\cite{dugan2024raid}}} & \multicolumn{4}{c}{\textbf{TSM~\citep{quaremba2026tsm}}} \\
\cmidrule(lr){2-5}\cmidrule(lr){6-9}\cmidrule(lr){10-13}\cmidrule(lr){14-17}
\textbf{Model} & \textbf{ArXiv} & \textbf{Reddit} & \textbf{Yelp} & \textbf{News} & \textbf{en} & \textbf{de} & \textbf{ru} & \textbf{zh} & \textbf{Cohere} & \textbf{GPT4} & \textbf{Llama} & \textbf{Mistral} & \textbf{FP} & \textbf{PE} & \textbf{SUM} & \textbf{TST} \\
\midrule
\rowcolor{gray!15} \multicolumn{17}{c}{\textbf{Zero-shot}} \\
\midrule
Likelihood & 0.6534 & 0.8620 & 0.8852 & 0.5793 & 0.7902 & 0.6935 & 0.6720 & 0.5848 & 0.7025 & 0.6937 & 0.7867 & 0.7339 & 0.5214 & 0.4344 & 0.5045 & 0.5096 \\
LLR & 0.7317 & 0.8226 & 0.8660 & 0.5965 & 0.7759 & 0.7130 & 0.6333 & 0.6500 & 0.7577 & 0.7212 & 0.8406 & 0.8091 & 0.5025 & 0.4130 & 0.5147 & 0.5107 \\
Rank & 0.5727 & 0.3702 & 0.3016 & 0.6061 & 0.2141 & 0.2942 & 0.3506 & 0.3846 & 0.3768 & 0.3217 & 0.2223 & 0.2600 & 0.5203 & 0.5561 & 0.4634 & 0.4886 \\
GECScore & 0.7848 & 0.5631 & 0.6283 & 0.7065 & 0.8057 & 0.6714 & 0.6133 & 0.4374 & 0.6655 & 0.4620 & 0.5497 & 0.6642 & 0.5459 & 0.5168 & 0.6678 & 0.5445 \\
Revise & 0.8527 & 0.7535 & 0.7917 & 0.7710 & 0.7740 & 0.6409 & 0.6835 & 0.6377 & 0.7364 & 0.6230 & 0.6676 & 0.7536 & 0.5726 & 0.5688 & 0.6261 & 0.5760 \\
RAIDAR & 0.8566 & 0.5395 & 0.7158 & 0.8490 & 0.8131 & 0.7943 & 0.6446 & 0.9153 & 0.8454 & 0.7234 & 0.7682 & 0.8444 & 0.6883 & 0.6706 & 0.7377 & 0.6729 \\
FastDetectGPT & 0.8563 & 0.6969 & 0.8304 & 0.6842 & 0.7850 & 0.7185 & 0.6500 & 0.6251 & 0.8922 & 0.7441 & 0.8221 & 0.7752 & 0.5060 & 0.4347 & 0.6082 & 0.4963 \\
Binoculars & 0.9393 & 0.9423 & 0.9661 & 0.8682 & 0.8082 & 0.7473 & 0.7118 & 0.7428 & 0.9312 & 0.8960 & 0.9490 & 0.9207 & 0.5668 & 0.4930 & 0.6294 & 0.5429 \\
\midrule
\rowcolor{gray!15} \multicolumn{17}{c}{\textbf{Supervised}} \\
\midrule
EditLens & 0.6732 & 0.4702 & 0.5050 & 0.3088 & 0.6769 & 0.7164 & 0.7510 & 0.6432 & 0.6532 & 0.5239 & 0.5835 & 0.5707 & 0.5840 & 0.6991 & 0.6092 & 0.6313 \\
ID & 0.7251 & 0.5616 & 0.6996 & 0.7681 & 0.5487 & 0.4998 & 0.5793 & 0.4900 & 0.5383 & 0.4832 & 0.5075 & 0.5280 & 0.5513 & 0.4715 & 0.5916 & 0.5312 \\
OpenAI-RoBERTa & 0.8357 & 0.7954 & 0.8101 & 0.7889 & 0.4683 & 0.3427 & 0.6985 & 0.4464 & 0.6855 & 0.5972 & 0.7070 & 0.7111 & 0.4483 & 0.5074 & 0.4309 & 0.5633 \\
RADAR & 0.9876 & 0.8733 & 0.9572 & 0.9976 & 0.6904 & 0.4682 & 0.3317 & 0.3392 & 0.8531 & 0.8689 & 0.8883 & 0.8896 & 0.5736 & 0.6302 & 0.4725 & 0.6192 \\
RepreGuard & 0.8960 & 0.9700 & 0.9780 & 0.9580 & 0.7060 & 0.6920 & 0.6400 & 0.7840 & 0.6020 & 0.7820 & 0.8920 & 0.8140 & 0.5040 & 0.5300 & 0.6040 & 0.4980 \\
BiScope & 0.9923 & 0.9820 & 0.9838 & 0.9925 & 0.8789 & 0.8014 & 0.7779 & 0.7904 & 0.9203 & 0.8791 & 0.9715 & 0.9481 & 0.7513 & 0.6011 & 0.7969 & 0.7056 \\
TextFluoroscopy & \cellcolor{orange!25}\underline{0.9948} & 0.9983 & 0.9951 & \cellcolor{orange!25}\underline{0.9980} & 0.9313 & 0.9015 & 0.8810 & 0.9032 & 0.8832 & 0.9546 & 0.9692 & 0.9492 & 0.7058 & 0.3723 & 0.2781 & 0.4089 \\
RoBERTa & \cellcolor{cyan!25}\textbf{1.0000} & 0.9996 & 0.9994 & \cellcolor{cyan!25}\textbf{1.0000} & 0.9507 & 0.8649 & 0.7762 & 0.8471 & 0.9404 & 0.9669 & 0.9770 & 0.9814 & 0.9361 & 0.7539 & 0.9333 & 0.7924 \\
\midrule
\rowcolor{gray!15} \multicolumn{17}{c}{\textbf{MGT Probes}} \\
\midrule
LLP & \cellcolor{cyan!25}\textbf{1.0000} & \cellcolor{cyan!25}\textbf{1.0000} & \cellcolor{cyan!25}\textbf{1.0000} & \cellcolor{cyan!25}\textbf{1.0000} & \cellcolor{cyan!25}\textbf{0.9565} & \cellcolor{cyan!25}\textbf{0.9337} & \cellcolor{cyan!25}\textbf{0.9212} & \cellcolor{orange!25}\underline{0.9443} & \cellcolor{cyan!25}\textbf{0.9714} & \cellcolor{cyan!25}\textbf{0.9925} & \cellcolor{orange!25}\underline{0.9946} & \cellcolor{cyan!25}\textbf{0.9928} & \cellcolor{orange!25}\underline{0.9707} & \cellcolor{cyan!25}\textbf{0.9424} & \cellcolor{cyan!25}\textbf{0.9783} & \cellcolor{cyan!25}\textbf{0.8960} \\
\hspace*{1em}$\Delta$ vs BL & \textcolor{green!60!black}{+0.00} & \textcolor{green!60!black}{+0.04} & \textcolor{green!60!black}{+0.06} & \textcolor{green!60!black}{+0.00} & \textcolor{green!60!black}{+0.58} & \textcolor{green!60!black}{+3.22} & \textcolor{green!60!black}{+4.02} & \textcolor{green!60!black}{+2.90} & \textcolor{green!60!black}{+3.09} & \textcolor{green!60!black}{+2.55} & \textcolor{green!60!black}{+1.76} & \textcolor{green!60!black}{+1.14} & \textcolor{green!60!black}{+3.46} & \textcolor{green!60!black}{+18.85} & \textcolor{green!60!black}{+4.50} & \textcolor{green!60!black}{+10.35} \\
CLP & \cellcolor{cyan!25}\textbf{1.0000} & \cellcolor{orange!25}\underline{0.9999} & \cellcolor{orange!25}\underline{0.9998} & \cellcolor{cyan!25}\textbf{1.0000} & \cellcolor{orange!25}\underline{0.9511} & \cellcolor{orange!25}\underline{0.9196} & \cellcolor{orange!25}\underline{0.9056} & \cellcolor{cyan!25}\textbf{0.9510} & \cellcolor{orange!25}\underline{0.9652} & \cellcolor{orange!25}\underline{0.9909} & \cellcolor{cyan!25}\textbf{0.9949} & \cellcolor{orange!25}\underline{0.9872} & \cellcolor{cyan!25}\textbf{0.9725} & \cellcolor{orange!25}\underline{0.9196} & \cellcolor{orange!25}\underline{0.9697} & \cellcolor{orange!25}\underline{0.8905} \\
\hspace*{1em}$\Delta$ vs BL & \textcolor{green!60!black}{+0.00} & \textcolor{green!60!black}{+0.04} & \textcolor{green!60!black}{+0.04} & \textcolor{green!60!black}{+0.00} & \textcolor{green!60!black}{+0.04} & \textcolor{green!60!black}{+1.81} & \textcolor{green!60!black}{+2.46} & \textcolor{green!60!black}{+3.57} & \textcolor{green!60!black}{+2.48} & \textcolor{green!60!black}{+2.39} & \textcolor{green!60!black}{+1.78} & \textcolor{green!60!black}{+0.58} & \textcolor{green!60!black}{+3.64} & \textcolor{green!60!black}{+16.57} & \textcolor{green!60!black}{+3.64} & \textcolor{green!60!black}{+9.80} \\
\bottomrule
\end{tabular}

    \end{adjustbox}}
    \caption{In-domain detection AUC scores across four benchmarks, comparing 16 baseline detectors (zero-shot and supervised) with MGT probes. 
    \colorbox{cyan!25}{\textbf{Bold}} indicates the best score per column, \colorbox{orange!25}{\underline{underline}} the second-best score, and ``$\Delta$ vs.\ BL'' the gain over the strongest baseline. LLP = Layer-Averaged Linear Probes; CLP = Concatenated-Layer Probes. For TSM, FP = First Paragraph, PE = Paragraph Extension, SUM = Summarization, and TST = Text Style Transfer. \textbf{Takeaway}: oth LLP and CLP consistently outperform all baselines by +0.04--18.85 AUC.}
    \label{tab:t_id}
\end{table*}

\paragraph{Concatenated-Layer Linear Probe (CLP)}
We also train a single linear probe on the concatenation of hidden states across layers. For each input $x_i$, we construct $\mathbf{h}_{i}^{\mathrm{concat}} =
\left[ \mathbf{h}_{i}^{(1)};\ldots;\mathbf{h}_{i}^{(L)}\right]\in \mathbb{R}^{L\times 100}$, which we use to train the same classifier as in Equation~\ref{eq:1} and perform inference analogously.
Compared to LLP, CLP learns a single MGT probe vector that \textit{jointly} identifies the most informative layers and the within-layer directions that encode MGT signals.

%%%%%%%%%%%%%%%%%%%%%%%%%%%%%%%%%%%%%%%%%%%%%%%%%%%%%%%%%%%%%%%%%%%%%%%%
% EXPERIMENTS
%%%%%%%%%%%%%%%%%%%%%%%%%%%%%%%%%%%%%%%%%%%%%%%%%%%%%%%%%%%%%%%%%%%%%%%%

\section{Experiments}
\label{sec:exp}

\subsection{Experimental Setup}

We provide full details on the benchmarks, detectors, and implementations in Appendix~\ref{app:exp_setup}.

\paragraph{Benchmarks}
We evaluate detectors on 4 benchmarks, each comprising 4 subsets that capture different dimensions of MGT.
We use DetectRL~\cite{wu2024detectrl} to study \textit{domains} (e.g., Reddit), 
MultiSocial~\cite{macko2025multi} to analyze \textit{languages} (e.g., Chinese),  
RAID~\cite{dugan2024raid} to different \textit{generators} (e.g., GPT4),
and TSM~\cite{quaremba2026tsm} to examine \textit{generation tasks} (e.g., summarization).
For each of the 16 subsets, we randomly sample 1,500 training instances and 500 test instances, balanced across labels.
Within each benchmark, the sampled data vary along additional dimensions (e.g., domains include multiple generators and adversarial attacks), resulting in more realistic and challenging detection settings.

\paragraph{Baselines}
We include 16 competitive MGT detectors.
For zero-shot detectors we include
\textit{Log-Likelihood}~\cite{solaiman2019loglikelihood}, 
\textit{LLR}~\cite{su2023detectllm}, 
\textit{Rank}~\cite{gehrmann2019gltr},  
\textit{GECScore}~\cite{wu2025wrote}, 
\textit{Revise}~\cite{zhu2023beat}, 
\textit{Raidar}~\cite{mao2024raidar}, 
\textit{FastDetectGPT}~\cite{bao2023fast}, 
and \textit{Binoculars}~\citep{hans2024binoculars}.

Fore supervised detectors we select 
\textit{EditLens}~\cite{thai2026editlens}, 
\textit{ID}~\cite{tulchinskii2023intrinsic}, 
\textit{OpenAI-RoBERTa}~\cite{solaiman2019loglikelihood},
\textit{RADAR}~\cite{hu2023radar}, 
\textit{RepreGuard}~\cite{chen2025repreguard},
\textit{BiScope}~\cite{guo2024biscope}, 
\textit{TextFluoroscopy}~\cite{yu2024fluo}, 
and fully fine-tune RoBERTa-base~\cite{liu2019roberta}.

\paragraph{Models and Metrics}
We implement probes with \texttt{Llama-3-8B}~\cite{grattafiori2024llama3} as the base model\footnote{We ablate model architectures an sizes in Appendix~\ref{app:ablation}.} and \texttt{sklearn} for the logistic regression with a $L2$ penalty of $C{=}1$. Following prior work~\cite{bao2023fast,hans2024binoculars} we report AUC as the evaluation metric.

% \begin{table*}[h]
%     \centering
%     {\renewcommand{\arraystretch}{1.0}
%     \begin{adjustbox}{width=\linewidth}
%         \input{tables/t_ood}
%     \end{adjustbox}}
%     \caption{OOD detection mean AUC scores across four benchmarks, comparing the four strongest detectors from Table~\ref{tab:t_id} against MGT probes. OOD columns report the average transfer performance across the three remaining subsets within each benchmark. Appendix Figure~\ref{fig:f_ood} presents the full OOD results. \textbf{Takeaway}: MGT probes exhibit strong OOD transferability, outperforming training-based detectors by 1.15--11.37 AUC points on average.}
%     \label{tab:t_ood}
% \end{table*}

\begin{table*}[h]
    \centering
    {\renewcommand{\arraystretch}{1.0}
    \begin{adjustbox}{width=1\linewidth}
        \begin{tabular}{lcccccccccccccccc}
\toprule
& \multicolumn{4}{c}{\textbf{DetectRL~\citep{wu2024detectrl}}} & \multicolumn{4}{c}{\textbf{MultiSocial~\citep{macko2025multi}}} & \multicolumn{4}{c}{\textbf{RAID~\cite{dugan2024raid}}} & \multicolumn{4}{c}{\textbf{TSM~\citep{quaremba2026tsm}}} \\
\cmidrule(lr){2-5}\cmidrule(lr){6-9}\cmidrule(lr){10-13}\cmidrule(lr){14-17}
\textbf{Model $\downarrow$ / OOD $\rightarrow$} & \textbf{ArXiv} & \textbf{Reddit} & \textbf{Yelp} & \textbf{News} & \textbf{en} & \textbf{de} & \textbf{ru} & \textbf{zh} & \textbf{Cohere} & \textbf{GPT4} & \textbf{Llama} & \textbf{Mistral} & \textbf{FP} & \textbf{PE} & \textbf{SUM} & \textbf{TST} \\
\midrule
TextFluoroscopy & 0.7580 & 0.9054 & 0.8691 & \cellcolor{orange!25}\underline{0.9402} & 0.8818 & 0.8423 & 0.8437 & \cellcolor{orange!25}\underline{0.8212} & 0.8402 & 0.9296 & 0.9236 & 0.9478 & 0.4052 & 0.5629 & 0.4807 & 0.4747 \\
BiScope & 0.8641 & 0.8426 & 0.9289 & 0.8545 & 0.7568 & 0.6565 & 0.6995 & 0.5169 & 0.9039 & 0.8482 & 0.9594 & 0.9294 & 0.6923 & 0.5744 & 0.6867 & 0.6692 \\
RepreGuard & 0.5327 & 0.5967 & 0.7687 & 0.5187 & 0.5047 & 0.5460 & 0.5020 & 0.5020 & 0.7047 & 0.7553 & 0.7367 & 0.7673 & 0.5007 & 0.5227 & 0.5793 & 0.4873 \\
RoBERTa & 0.9108 & 0.8735 & 0.8491 & 0.7781 & 0.8511 & 0.7870 & 0.7784 & 0.6389 & 0.9118 & 0.9554 & 0.9688 & 0.9503 & 0.8324 & 0.7367 & 0.8452 & 0.7376 \\
\midrule
LLP & \cellcolor{cyan!25}\textbf{0.9830} & \cellcolor{cyan!25}\textbf{0.9800} & \cellcolor{cyan!25}\textbf{0.9897} & \cellcolor{cyan!25}\textbf{0.9441} & \cellcolor{cyan!25}\textbf{0.9321} & \cellcolor{cyan!25}\textbf{0.8820} & \cellcolor{cyan!25}\textbf{0.8945} & \cellcolor{cyan!25}\textbf{0.8696} & \cellcolor{cyan!25}\textbf{0.9267} & \cellcolor{cyan!25}\textbf{0.9812} & \cellcolor{cyan!25}\textbf{0.9908} & \cellcolor{cyan!25}\textbf{0.9821} & \cellcolor{cyan!25}\textbf{0.9513} & \cellcolor{cyan!25}\textbf{0.8087} & \cellcolor{cyan!25}\textbf{0.9512} & \cellcolor{cyan!25}\textbf{0.8560} \\
\hspace*{1em}$\Delta$ vs BL & \textcolor{green!60!black}{+7.22} & \textcolor{green!60!black}{+7.46} & \textcolor{green!60!black}{+6.08} & \textcolor{green!60!black}{+0.39} & \textcolor{green!60!black}{+5.03} & \textcolor{green!60!black}{+3.96} & \textcolor{green!60!black}{+5.08} & \textcolor{green!60!black}{+4.84} & \textcolor{green!60!black}{+1.49} & \textcolor{green!60!black}{+2.58} & \textcolor{green!60!black}{+2.20} & \textcolor{green!60!black}{+3.17} & \textcolor{green!60!black}{+11.89} & \textcolor{green!60!black}{+7.20} & \textcolor{green!60!black}{+10.60} & \textcolor{green!60!black}{+11.83} \\
CLP & \cellcolor{orange!25}\underline{0.9807} & \cellcolor{orange!25}\underline{0.9665} & \cellcolor{orange!25}\underline{0.9867} & 0.9263 & \cellcolor{orange!25}\underline{0.9087} & \cellcolor{orange!25}\underline{0.8651} & \cellcolor{orange!25}\underline{0.8759} & 0.8043 & \cellcolor{orange!25}\underline{0.9200} & \cellcolor{orange!25}\underline{0.9723} & \cellcolor{orange!25}\underline{0.9839} & \cellcolor{orange!25}\underline{0.9791} & \cellcolor{orange!25}\underline{0.9317} & \cellcolor{orange!25}\underline{0.7943} & \cellcolor{orange!25}\underline{0.9152} & \cellcolor{orange!25}\underline{0.8451} \\
\hspace*{1em}$\Delta$ vs BL & \textcolor{green!60!black}{+6.99} & \textcolor{green!60!black}{+6.11} & \textcolor{green!60!black}{+5.78} & \textcolor{orange!85!black}{-1.40} & \textcolor{green!60!black}{+2.69} & \textcolor{green!60!black}{+2.27} & \textcolor{green!60!black}{+3.22} & \textcolor{orange!85!black}{-1.69} & \textcolor{green!60!black}{+0.82} & \textcolor{green!60!black}{+1.69} & \textcolor{green!60!black}{+1.51} & \textcolor{green!60!black}{+2.88} & \textcolor{green!60!black}{+9.93} & \textcolor{green!60!black}{+5.76} & \textcolor{green!60!black}{+7.01} & \textcolor{green!60!black}{+10.75} \\
\bottomrule
\end{tabular}

    \end{adjustbox}}
    \caption{OOD detection mean AUC scores across four benchmarks, comparing the four strongest detectors from Table~\ref{tab:t_id} against MGT probes. OOD columns report the average transfer performance across the three remaining subsets within each benchmark (e.g., the ArXiv column averages Reddit $\rightarrow$ ArXiv, Yelp $\rightarrow$ ArXiv, and News $\rightarrow$ ArXiv).~\colorbox{cyan!25}{\textbf{Bold}} indicates the best score per column, \colorbox{orange!25}{\underline{underline}} the second-best score, and ``$\Delta$ vs.\ BL'' the gain over the strongest baseline. Appendix Figure~\ref{fig:f_ood} presents the full OOD results. \textbf{Takeaway}: MGT probes exhibit strong OOD transferability, outperforming training-based detectors by +0.39--11.37 AUC on average.}
    \label{tab:t_ood}
\end{table*}

% \begin{table*}[h]
%     \centering
%     {\renewcommand{\arraystretch}{1.0}
%     \begin{adjustbox}{width=.9\linewidth}
%         \input{tables/t_ood}
%     \end{adjustbox}}
%     \caption{OOD detection mean AUC scores across four benchmarks, comparing the four strongest detectors from Table~\ref{tab:t_id} against MGT probes. OOD columns report the average transfer performance across the three remaining subsets within each benchmark. Appendix Figure~\ref{fig:f_ood} presents the full OOD results. \textbf{Takeaway}: MGT probes exhibit strong OOD transferability, outperforming training-based detectors by 0.39--11.37 AUC on average.}
%     \label{tab:t_ood}
% \end{table*}

%%%%%%%%%%%%%%%%%%%%%%%%%%%%%%%%%%%%%%%%%%%%%%%%%%%%%%%%%%%%%%%%%%%%%%%%
% ID
%%%%%%%%%%%%%%%%%%%%%%%%%%%%%%%%%%%%%%%%%%%%%%%%%%%%%%%%%%%%%%%%%%%%%%%%
\subsection{In-domain (ID) Detection}
\label{sec:exp:id}

Our findings in Section~\ref{sec:mgt_rep} suggest that simple linear probes may provide effective detection, we therefore begin by asking \textbf{RQ1}: \textit{How do MGT probes compare to MGT detectors in-domain?}

\textbf{MGT probes consistently outperform baselines across 16 in-domain settings by 0.04--18.85 AUC.}
Table~\ref{tab:t_id} reports AUC results across 16 ID settings, comparing 16 detectors with our MGT probes.
Both variants consistently outperform baselines, achieving gains of up to 18.85 AUC relative to the strongest detector.
The near-identical performance of CLP and LLP suggests that a single probe trained on concatenated layer representations is sufficient for ID detection.

On DetectRL, baseline detectors exhibit near-saturated performance despite the domain subsets containing text generated by multiple LLMs and subjected to diverse adversarial attacks (e.g., paraphrasing).
As the benchmark provides little headroom for improvement, probes achieve only relatively limited gains.\footnote{We also experimented with domains from RAID and M4GT, observing similar saturation effects.}
On MultiSocial, probes achieve the largest improvements over baselines on non-English languages (+1.81--4.02 AUC), suggesting that the learned probing directions are language-agnostic, for which we find support in OOD (\S\ref{sec:exp:ood}) and probing vector experiments (\S\ref{sec:exp:probes}).
Across generators on RAID, probes show additional robust gains over already strong baselines (+0.58--3.09 AUC).
TSM presents the most challenging detection setting, where even supervised detectors struggle with mixed human--machine text (PE) and shorter texts containing only minor stylistic edits (TST). Nevertheless, MGT probes maintain strong performance, achieving the largest gains over baselines (+3.46--18.85 AUC).

% Among zero-shot methods, RepreGuard and FastDetectGPT achieve competitive performance on DetectRL and RAID, but deteriorate substantially on TSM and MultiSocial, highlighting the limitations of zero-shot approaches on more challenging benchmarks.
% Supervised baselines generally perform more strongly and consistently, particularly BiScope, TextFluoroscopy, and RoBERTa.
% However, both MGT probe variants consistently outperform all baselines, with the largest improvements observed on MultiSocial (+1.22 to +3.46 AUC points).

% These results indicate that the linear separability of HWT and MGT in activation space persists even under challenging conditions, including shorter and more informal multilingual texts (MultiSocial), as well as diverse text-generation tasks (TSM).

% Notably, LLP and CLP perform almost identically across benchmarks.
% Although CLP slightly improves performance on some multilingual settings, LLP often marginally outperforms ($<$ 1 AUC) CLP on TSM and RAID.
% Overall, our findings provide strong evidence that simple MGT probes are sufficient for strong ID detection performance, corroborating that MGT signals are linearly encoded throughout model representations.

% \hl{to do} it may make sense to first do the probe comparison, and from there hypothesis a "universal" direction which leads to higher OOD; then confirm this in experiments

%%%%%%%%%%%%%%%%%%%%%%%%%%%%%%%%%%%%%%%%%%%%%%%%%%%%%%%%%%%%%%%%%%%%%%%%
% OOD
%%%%%%%%%%%%%%%%%%%%%%%%%%%%%%%%%%%%%%%%%%%%%%%%%%%%%%%%%%%%%%%%%%%%%%%%

\subsection{Out-of-domain (OOD) Detection}
\label{sec:exp:ood}

If probes recover genuine MGT signals rather than dataset artifacts---a key limitation of existing supervised detectors~\cite{doughman2025exploring}-- they should transfer OOD.
\textbf{RQ2}: \textit{How well do MGT probes compare to existing detectors in OOD settings?}

\textbf{MGT probes preserve strong OOD transferability, outperforming training-based detectors by 0.39--11.37 AUC.}
Table~\ref{tab:t_ood} reports mean OOD AUC for each target subset not seen during training. Each column averages transfer performance from the three remaining subsets within the same benchmark. For example, the ArXiv column averages Reddit $\rightarrow$ ArXiv, Yelp $\rightarrow$ ArXiv, and News $\rightarrow$ ArXiv.\footnote{Appendix Figure~\ref{fig:f_ood} reports the full OOD transfer matrix.}
As baselines, we include the strongest training-based detectors from Table~\ref{tab:t_id}.
Overall, MGT probes transfer effectively across benchmark settings, improving over baselines by up to 11.83 AUC. LLP consistently exhibits more robust transfer performance than CLP.

On DetectRL, LLP exhibits strong OOD transferability across domains, outperforming baselines by 0.39--7.46 AUC. 
CLP achieves similarly strong transfer except on News, which we attribute to overfitting to stylistic cues associated with formal writing.\footnote{We further analyze CLP's transfer behavior on News and Chinese in Appendix~\ref{app:ood}.}
On MultiSocial, LLP demonstrates consistent and reliable cross-lingual transfer, even between typologically distant languages (e.g., ru$\rightarrow$zh). 
We interpret this finding as evidence for a language-agnostic MGT direction encoded across layers. 
In contrast, CLP exhibits weaker cross-lingual transfer and fails to generalize from ru$\rightarrow$zh, suggesting that its globally learned representation is less robust for OOD transfer.
As in the ID setting, baselines already perform strongly on RAID's cross-generator transfer.
Nevertheless, both probes achieve modest but consistent improvements.

The largest gains occur in cross-task transfer, where probes outperform baselines by 5.76--11.89 AUC. 
Similar to the ID setting, supervised detectors struggle to generalize across tasks, particularly on PE and TST. Despite these challenges, both probes maintain strong performance, achieving AUC ${>}$0.80 in these difficult settings.

%%%%%%%%%%%%%%%%%%%%%
% Disucssion
%%%%%%%%%%%%%%%%%%%%%
% How to these generalizability findings compare to prior results where papers used LPs for OOD?

    % \cite{tan2024analysing}

    % - https://arxiv.org/pdf/2505.22637v1

    % - And google paper

    % - https://openreview.net/forum?id=H5sbfvEbTh

% Conclude with: We find compelling evidence that simple MGT probes are more robust than existing training-based detectors.

%%%%%%%%%%%%%%%%%%%%%%%%%%%%%%%%%%%%%%%%%%%%%%%%%%%%%%%%%%%%%%%%%%%%%%%%
% OOB
%%%%%%%%%%%%%%%%%%%%%%%%%%%%%%%%%%%%%%%%%%%%%%%%%%%%%%%%%%%%%%%%%%%%%%%%
% \subsection{Out-of-benchmark Detection}

% \textit{Does probing identify a linear MGT direction that transfers across benchmarks?}

% Idea: argue that often results on one benchmark do not transfer to another; indicating that models overfit or overrely on the characteristics of one benchmark; is this mini-benchmark iclr paper useful maybe?

%%%%%%%%%%%%%%%%%%%%%%%%%%%%%%%%%%%%%%%%%%%%%%%%%%%%%%%%%%%%%%%%%%%%%%%%
% LAYER ANALYSIS
%%%%%%%%%%%%%%%%%%%%%%%%%%%%%%%%%%%%%%%%%%%%%%%%%%%%%%%%%%%%%%%%%%%%%%%%
\subsection{Probing Vector Similarity}
\label{sec:exp:probes}

\begin{figure}[h]
     \centering
     \includegraphics[width=.8\linewidth]{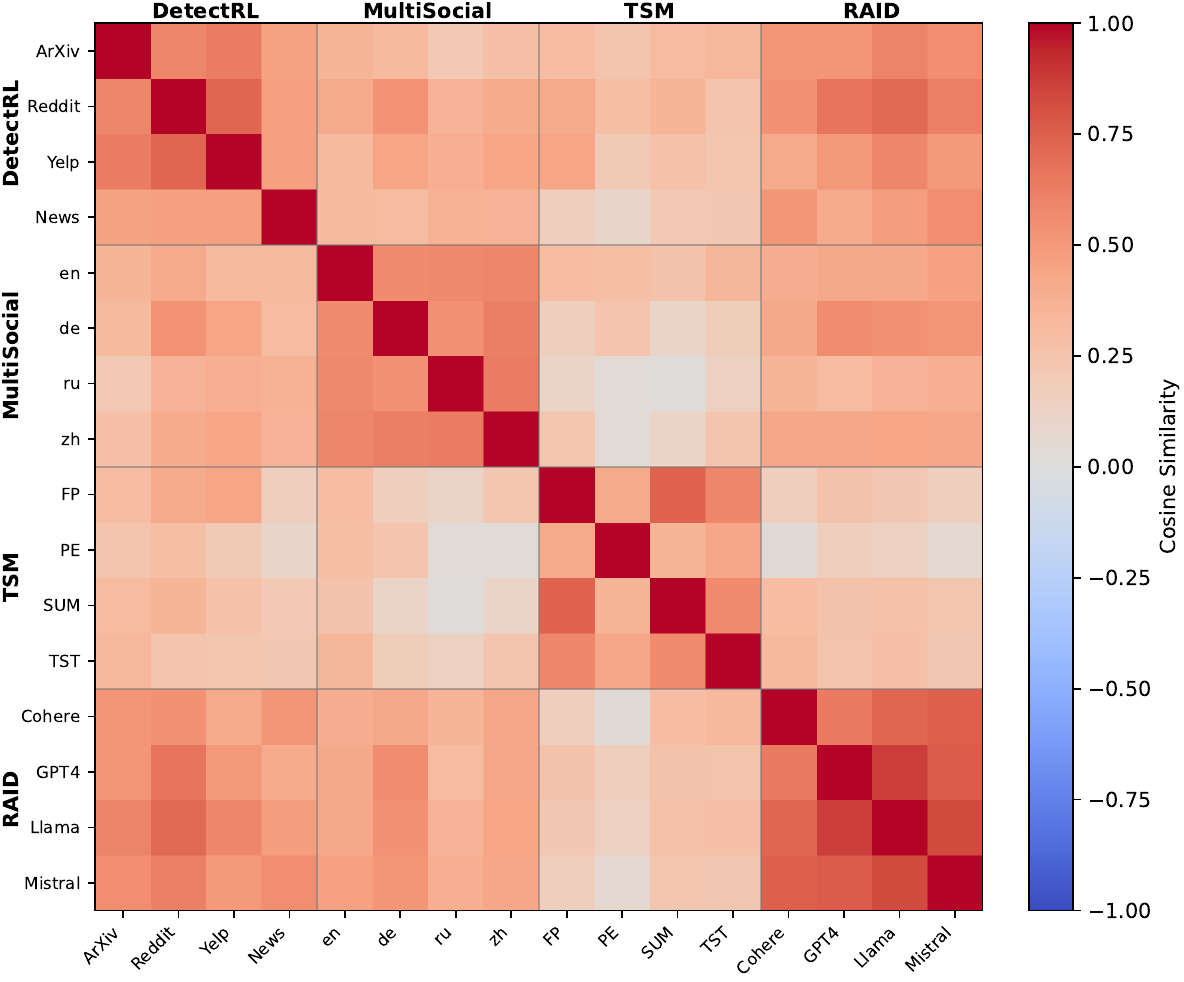}
    \caption{Cosine similarities of probing vectors at layer 16 across all 16 datasets. \textbf{Takeaway}: Probing vectors exhibit high similarity both within and across datasets, suggesting the existence of a shared latent MGT direction.}
    \label{fig:f_probes_pca}
\end{figure}

To better understand the strong transferability of probes, we ask
\textbf{RQ3}: \textit{Why do MGT probes generalize across OOD settings?}
Prior work has identified \textit{universal} probing directions that can be recovered across settings~\cite{agarwal2025context,wang2026refusal}.
We therefore analyze the alignment of MGT probing vectors at the layer with the highest OOD AUC performance (see Appendix Figure~\ref{fig:f_layer}).

\textbf{Probing vectors exhibit high within-benchmark and moderate cross-benchmark similarity, suggesting a shared latent MGT direction.}
Figure~\ref{fig:f_probes_pca} shows the pairwise cosine similarities of LLP probing vectors at layer 16 across all 16 evaluation subsets.
We find that probing vectors are highly aligned within benchmarks and moderately aligned across benchmarks.
This suggests that probes recover a relatively stable latent MGT direction, helping explain the strong OOD transfer observed in Table~\ref{tab:t_ood}.
Additionally, Appendix Figure~\ref{fig:f_drl_parallel} illustrates that latent human-machine directions are approximately parallel across subsets of DetectRL. 
Together, ``machineness'' appears to be encoded as a shared and therefore transferable linear direction in latent space.
%%%%%%%%%%%%%%%%%%%%%
% Disucssion
%%%%%%%%%%%%%%%%%%%%%
% See heatmaps in Why context matters

%%%%%%%%%%%%%%%%%%%%%%%%%%%%%%%%%%%%%%%%%%%%%%%%%%%%%%%%%%%%%%%%%%%%%%%%
    % SAMPLE-EFFICIENCY ANALYSIS
%%%%%%%%%%%%%%%%%%%%%%%%%%%%%%%%%%%%%%%%%%%%%%%%%%%%%%%%%%%%%%%%%%%%%%%%
\subsection{Sample-efficiency Analysis}
\label{sec:exp:sample}

\begin{figure}[h]
     \centering
     \includegraphics[width=\linewidth]{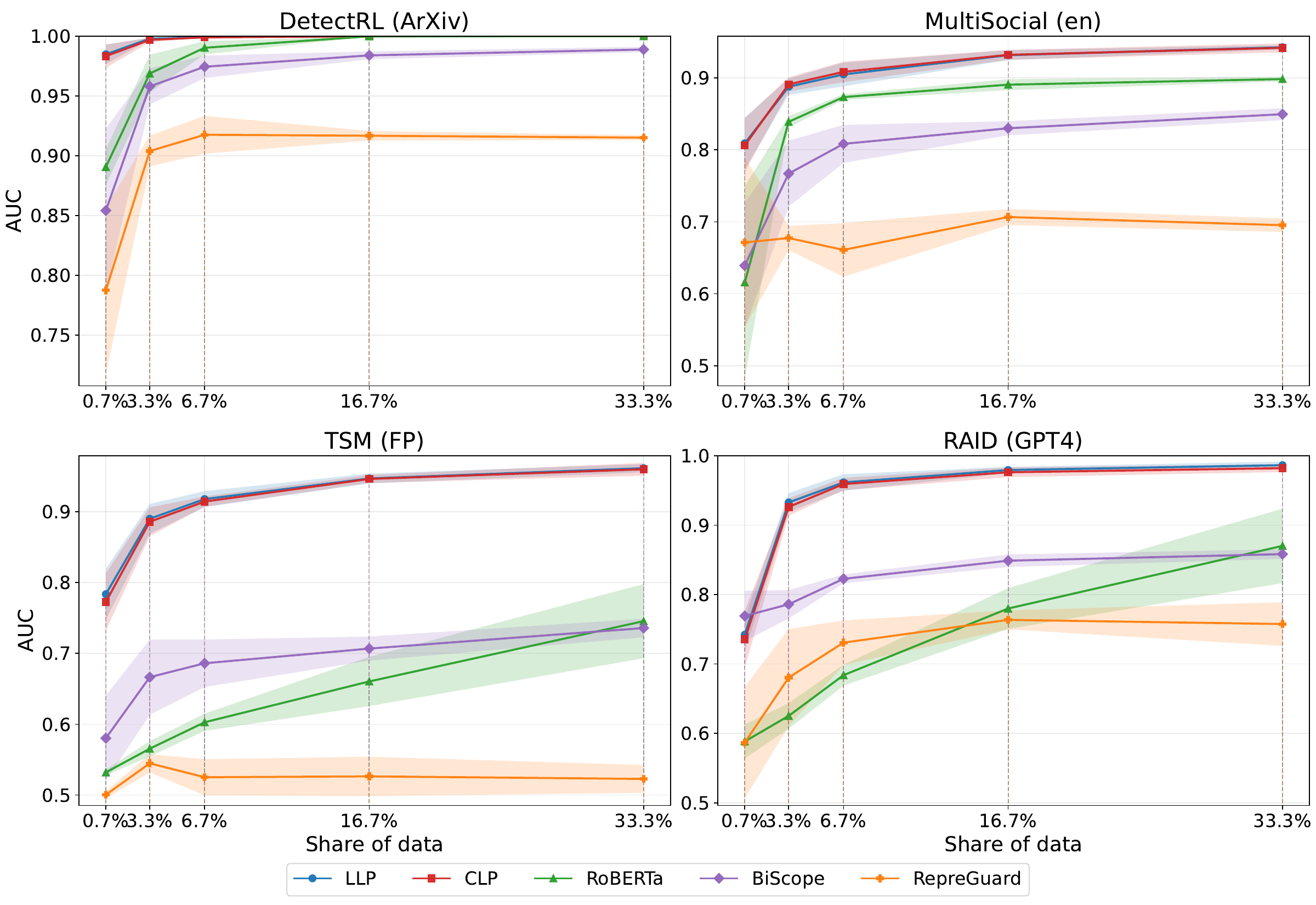}
    \caption{In-domain AUC scores for selected datasets as a function of training set size. \textbf{Takeaway}: MGT probes achieve strong performance with only 10--100 training samples.}
    \label{fig:f_samples_pca}
\end{figure}

Given the probes' simplicity and the low-dimensional, linear MGT signal, we expect them to require fewer labeled examples than more complex detectors, hence
\textbf{RQ4}: \textit{How sample-efficient are MGT probes compared to existing MGT detectors?}

\textbf{MGT probes achieve strong performance with only 10--100 training samples while exhibiting low sampling uncertainty.}
Figure~\ref{fig:f_samples_pca} plots AUC as a function of training set size for one subset from each benchmark.\footnote{We still fit the PCA subspace for each layer using the full training set. Appendix Figure~\ref{app:fig:f_samples} shows nearly identical results without PCA-reduced activations.}
To quantify sampling uncertainty, we repeat each experiment with different random seeds.
Compared to the strongest baseline, RoBERTa, both probes improve rapidly, reaching near-peak performance with only 3.3--6.7\% of the training data (10--100 samples) before plateauing.
Moreover, probes exhibit substantially lower sampling uncertainty than RepreGuard and RoBERTa, indicating that the latent MGT direction can be recovered reliably even in low-resource settings.

%%%%%%%%%%%%%%%%%%%%%
% Disucssion
%%%%%%%%%%%%%%%%%%%%%
% Find and discuss sample efficiency of other papers

% Same sample-efficiency: https://arxiv.org/pdf/2507.12428

%%%%%%%%%%%%%%%%%%%%%%%%%%%%%%%%%%%%%%%%%%%%%%%%%%%%%%%%%%%%%%%%%%%%%%%%
% AI-EDITING
%%%%%%%%%%%%%%%%%%%%%%%%%%%%%%%%%%%%%%%%%%%%%%%%%%%%%%%%%%%%%%%%%%%%%%%%
\subsection{Detecting varying degrees of AI-editing}
\label{sec:exp:edit}

\begin{figure}[h]
     \centering
     \begin{subfigure}[b]{1\linewidth}
         \centering
         \includegraphics[width=1\linewidth]{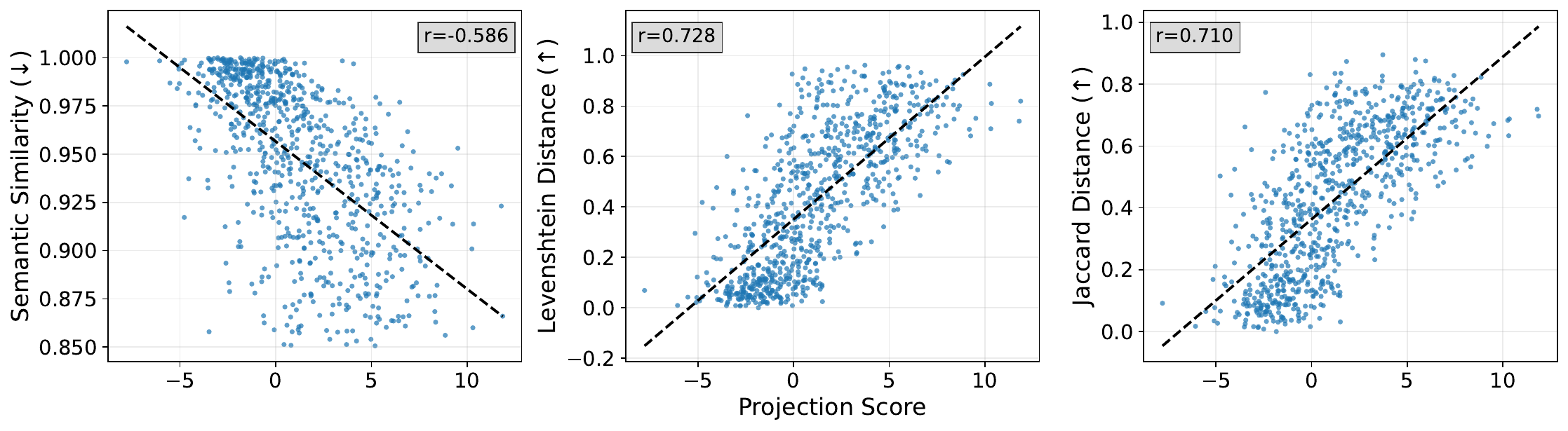}
         \caption{LLP}
     \end{subfigure}
     
      \vfill \vspace{0.3cm}
     
     \begin{subfigure}[b]{1\linewidth}
         \centering
         \includegraphics[width=1\linewidth]{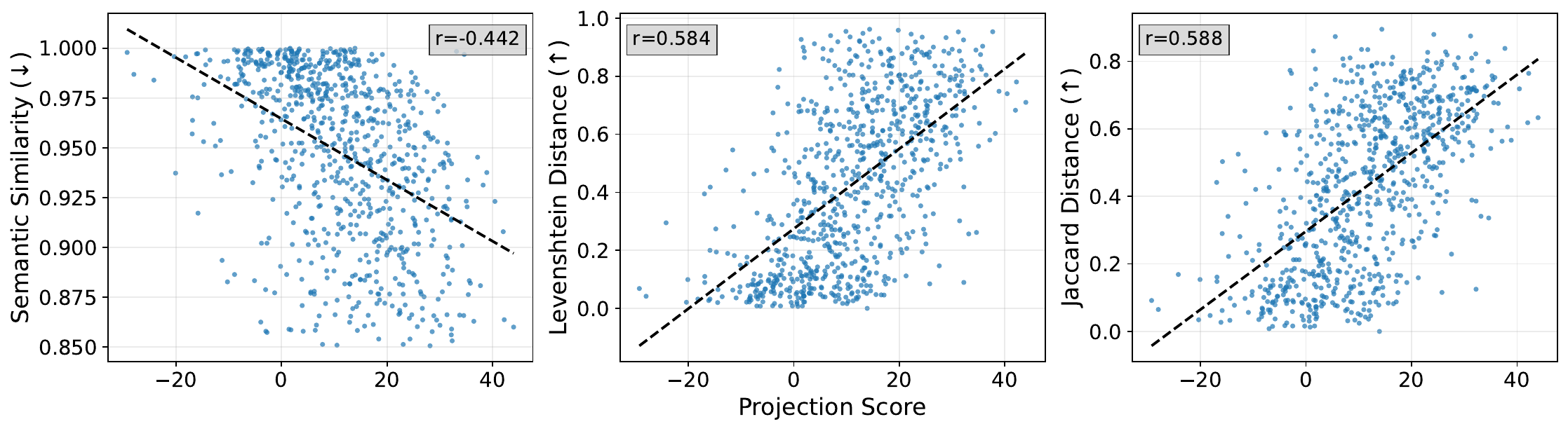}
         \caption{CLP}
     \end{subfigure} 
    \caption{Edit-strength metrics for samples from APT-Eval~\cite{saha2025apt} plotted against probe projection scores. The dashed line shows the linear regression fit, and $r$ denotes the Pearson correlation coefficient. \textbf{Takeaway}: Projection scores correlate strongly with edit-strength metrics, suggesting that the latent MGT direction captures a continuous spectrum of ``machineness.''}
    \label{fig:f_edit_apt}
    % \vspace{-1em}
\end{figure}

% Old table replaced by figure

% \begin{table}[H]

%     \centering
%     \begin{adjustbox}{width=.7\linewidth}
%         \input{tables/t_edits}
%     \end{adjustbox}
%     \caption{Pearson correlation coefficients between probe projection scores and text-similarity metrics for AI-edited text with varying degrees of editing. \textbf{Takeaway}: Across both datasets, LLP projection scores show strong correlations, indicating that the probes encode fine-grained levels of AI-edited text.}
%     \label{tab:t_edits}
% \end{table}

While the previous experiments focus on binary detection, a growing body of work seeks to identify human-written text that has been edited by AI~\cite{zhang2024llm}.
Probe projection scores provide a natural continuous measure of MGT, which raises
\textbf{RQ5}: \textit{Do MGT probes encode fine-grained signals of AI-polished text?}
To test this hypothesis, we use APT-Eval~\cite{saha2025apt} and EditLens~\cite{thai2026editlens}, which contain AI-edited texts with varying degrees of edits.
We correlate probe projection scores with three metrics that quantify the degree of AI editing (cosine similarity, Levenshtein and Jaccard distance) as in~\citet{saha2025apt}.

\textbf{Probe projection scores strongly correlate with the degree of AI editing, indicating that probing vectors capture fine-grained levels of AI-polished text.}
Figure~\ref{fig:f_edit_apt} plots the text-similarity metrics against projection scores for LLP (top) and CLP (bottom).
The dashed line shows a linear regression fit, and $r$ denotes the Pearson correlation coefficient.
LLP exhibits strong correlations across all similarity metrics, indicating that human text subjected to stronger AI editing is projected closer to the machine-text region of latent space.
Crucially, this behavior emerges despite the probes being trained only with binary labels.
In contrast, CLP shows only moderate correlations, which we attribute to the difficulty of capturing subtle AI edits in a joint space.
These findings suggest that the latent MGT direction represents a \textit{continuous spectrum} of ``machineness,'' enabling fine-grained MGT detection beyond binary classification.
Appendix Figure~\ref{fig:f_edit_editlens} shows similar results for EditLens.

%%%%%%%%%%%%%%%%%%%%%
% Disucssion
%%%%%%%%%%%%%%%%%%%%%
% How to these generalizability findings compare to prior results where papers used LPs for OOD?

    % Multi layer ensemble in the paper LP scale with size; they also show that combining multiple layers yield better performance; we show the same 

%%%%%%%%%%%%%%%%%%%%%%%%%%%%%%%%%%%%%%%%%%%%%%%%%%%%%%%%%%%%%%%%%%%%%%%%
% CONCLUSION
%%%%%%%%%%%%%%%%%%%%%%%%%%%%%%%%%%%%%%%%%%%%%%%%%%%%%%%%%%%%%%%%%%%%%%%%
\section{Conclusion}

In this work, we provide empirical evidence that machine-generated text (MGT) and human-written text occupy linearly separable regions in latent space, and offer a potential explanation through analyses of representation quality metrics. We further show that simple MGT probes---linear probes trained on LM hidden-state activations---achieve stronger OOD robustness while requiring fewer training samples than supervised detectors. Finally, we show that the continuous probing direction captures fine-grained levels of AI editing.

Overall, our findings suggest that ``machineness'' is encoded as a stable, continuous, and linearly accessible direction in activation space, enabling both robust detection and fine-grained estimation of AI-edited text. Future work could explore alternative probing variants, investigate how these representations change under adversarial attacks, and further extend the approach to AI-edited text detection.

% Our findings suggest that ``machineness'' is encoded as a stable, continuous, and linearly accessible direction in activation space, enabling robust MGT detection and fine-grained estimation of AI editing.

%%%%%%%%%%%%%%%%%%%%%%%%%%%%%%%%%%%%%%%%%%%%%%%%%%%%%%%%%%%%%%%%%%%%%%%%
% LIMITATIONS
%%%%%%%%%%%%%%%%%%%%%%%%%%%%%%%%%%%%%%%%%%%%%%%%%%%%%%%%%%%%%%%%%%%%%%%%
\section*{Limitations}

\paragraph{Non-Exhaustive Analysis of Representational Differences}
We analyze HWT and MGT representations using four representation-quality metrics.
However, these metrics capture only selected information-theoretic and geometric properties of latent representations, and therefore provide a non-exhaustive view of their distributional differences.
While the observed differences help explain the existence of a linear decision boundary, other important representational properties may also contribute to the separation between HWT and MGT.
Nevertheless, our analysis offers an initial characterization of how human- and machine-written text differ in latent space.

\paragraph{Model Selection}
Although we ablate probe performance across model architectures and sizes, we do not systematically study how detection performance scales with model size or architectural design choices.
Future work could investigate these factors in greater depth and analyze how they influence the emergence and transferability of MGT signals.

\paragraph{Linear Probing Variants}
We focus on the simplest probe architecture: logistic regression.
Prior work has proposed more expressive alternatives, including attention-based probes and learned pooling strategies, which may recover richer MGT signals.
Exploring such probe variants represents a promising direction for future research.

\paragraph{Universality of the Latent MGT Direction}
Although our results show that MGT and HWT are linearly separable and that linear probes transfer effectively across OOD settings, they do not establish the existence of a \textit{universal} latent MGT direction.
Representation spaces contain many potential confounding factors, and probing provides only one method for identifying latent concepts.
Future work should further investigate the stability, causality, and universality of MGT directions across models and data distributions.

\paragraph{Fine-Grained Detection of AI-Edited Text}
While we show that probe projection scores correlate strongly with the degree of AI editing, we do not evaluate probes as dedicated fine-grained detectors or compare them against specialized multiclass or regression-based approaches.
Our goal is to establish that the latent MGT direction encodes meaningful variation in AI involvement.
Future work could optimize probe training for continuous prediction and benchmark their competitiveness on fine-grained AI-editing tasks.

\paragraph{Limited applicability to accessible models}
Linear probes~\cite{alain2017understanding} require access to model internals. A limitation of MGT probes is therefore that they are only applicable to open-source models for which hidden states are accessible.

%%%%%%%%%%%%%%%%%%%%%%%%%%%%%%%%%%%%%%%%%%%%%%%%%%%%%%%%%%%%%%%%%%%%%%%%
% LIMITATIONS
%%%%%%%%%%%%%%%%%%%%%%%%%%%%%%%%%%%%%%%%%%%%%%%%%%%%%%%%%%%%%%%%%%%%%%%%
\section*{Ethical Considerations}

We do not identify any risks arising from this work.
We use the AI assistant ChatGPT for proofreading and for assistance with formatting tables and figures.

\section*{Acknowledgements}

This work was supported by the Engineering and Physical Sciences Research Council [grant number Y009800/1], through funding from Responsible AI UK (KP0011), as part of the Participatory Harm Auditing Workbenches and Methodologies (\href{https://phawm.org/}{PHAWM}) project, and by UK Research and Innovation [grant number EP/S023356/1], through the UKRI Centre for Doctoral Training in Safe and Trusted Artificial Intelligence (\href{https://safeandtrustedai.org/}{www.safeandtrustedai.org}).

\clearpage

% Custom bibliography entries only
\bibliography{custom}

\clearpage
\clearpage

\appendix

% \onecolumn

%%%%%%%%%%%%%%%%%%%%%%%%%%%%%%%%%%%%%%%%%%%%%%%%%%%%%%%%%%%%%%%%%%%%%%%%
% MGT REPRESENTATIONS
%%%%%%%%%%%%%%%%%%%%%%%%%%%%%%%%%%%%%%%%%%%%%%%%%%%%%%%%%%%%%%%%%%%%%%%%
\section{MGT Representation Analysis}
\label{app:mgt_rep}

\subsection{Representation Quality Metrics}
\label{app:mgt_rep:metrics}

We briefly introduce the representation-quality metrics used in our analysis and provide their formulae below.
Following~\citet{skean2025layer}, we group these metrics into \textit{information-theoretic}, \textit{geometric}, and \textit{invariance-based} categories.
Our analysis focuses on the first two categories, which characterize the complexity, dimensionality, and geometry of MGT and HWT representations.

\begin{enumerate}
    \item \textbf{Entropy}~\cite{giraldo2015entropy} measures how much information a representation contains.
    Higher values indicate more diverse and information-rich features, whereas lower values suggest more redundant and compressed representations. 
    For activations at layer $l$, let $Z^{(l)} \in \mathbb{R}^{d \times N}$ denote the representation matrix and define the corresponding Gram matrix as:

    \[
    K^{(l)} = Z^{(l)} {Z^{(l)}}^{\top}.
    \]
    
    Let $\{\lambda_i(K^{(l)})\}$ denote the non-negative eigenvalues of $K^{(l)}$.
    For any order $\alpha > 0$, the entropy is defined as:
    
    \[
\operatorname{Entropy}(Z^{(l)})
    =
    \frac{1}{1-\alpha}
    \log
    \sum_{i=1}^{r}
    \left(
    \frac{\lambda_i(K^{(l)})}{\operatorname{tr}(K^{(l)})}
    \right)^{\alpha},
    \]
    
    where $r = \operatorname{rank}(K^{(l)}) \leq \min(N, d)$.
    Following~\citet{skean2025layer}, we compute the von Neumann entropy by setting $\alpha = 1$.

    \item \textbf{Effective Rank}~\cite{roy2007effective,wei2024differank} measures the dimensional complexity of representations.
    Higher values indicate more distributed and noisy feature representations, while lower values suggest stronger compression and more compact representations.
    For activations at layer $l$, represented by the matrix $A^{(l)} \in \mathbb{R}^{d \times N}$ with $N$ samples and hidden dimension $d$, the effective rank is defined as:
    
    \[
    \operatorname{eR}(A^{(l)})
    =
    \exp\left(
    -
    \sum_{i=1}^{Q}
    \frac{\sigma_i}{\sum_{j=1}^{Q} \sigma_j}
    \log
    \frac{\sigma_i}{\sum_{j=1}^{Q} \sigma_j}
    \right),
    \]
    
    where $Q = \min(N, d)$ and $\sigma_1, \dots, \sigma_Q$ denote the singular values of $A^{(l)}$.
    
    \item \textbf{Anisotropy}~\cite{razzhigaev2024shape} measures the degree to which activations concentrate along specific directions in latent space.
    Higher anisotropy indicates that activations are strongly aligned with a smaller set of dominant directions, whereas lower values indicate more uniformly distributed representations.
    To compute anisotropy, let $X^{(l)} \in \mathbb{R}^{N \times d}$ denote the centered activations at layer $l$, where $\sigma_1, \dots, \sigma_k$ are the singular values of $X^{(l)}$.
    The anisotropy score is defined as:
    
    \[
    \operatorname{Anisotropy}(X^{(l)})
    =
    \frac{\sigma_1^2}{\sum_{i=1}^{k} \sigma_i^2},
    \]
    
    where $k = \min(N, d)$.

    \item \textbf{Intrinsic Dimension}~\cite{Facco2017estimating} estimates the minimum number of dimensions required to describe the local structure of representations without substantial information loss.
    Higher values indicate richer and more complex latent structures, whereas lower values suggest that representations lie on simpler and lower-dimensional manifolds.
    We estimate intrinsic dimensionality using the Two-Nearest Neighbor estimator proposed by~\citet{Facco2017estimating}, implemented via the \texttt{scikit-dimension} package.\footnote{\url{https://github.com/scikit-learn-contrib/scikit-dimension}}
    We refer readers to the original work for details on the estimator and its formula.
        
\end{enumerate}

\subsection{Extended Analysis}
\label{app:mgt_rep:results}

\begin{figure}[h]
    \centering
    % Reddit PCA
    \begin{subfigure}[b]{1\linewidth}
        \centering
        \includegraphics[width=\linewidth]{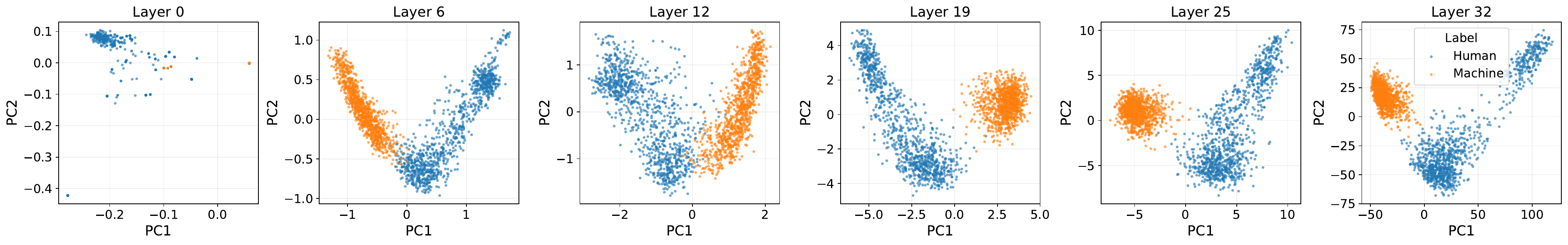}
        \caption{Reddit}
        \label{fig:pca_reddit}
    \end{subfigure}
    
    \vspace{0.3cm}
    
    % PeerRead PCA
    \begin{subfigure}[b]{1\linewidth}
        \centering
        \includegraphics[width=\linewidth]{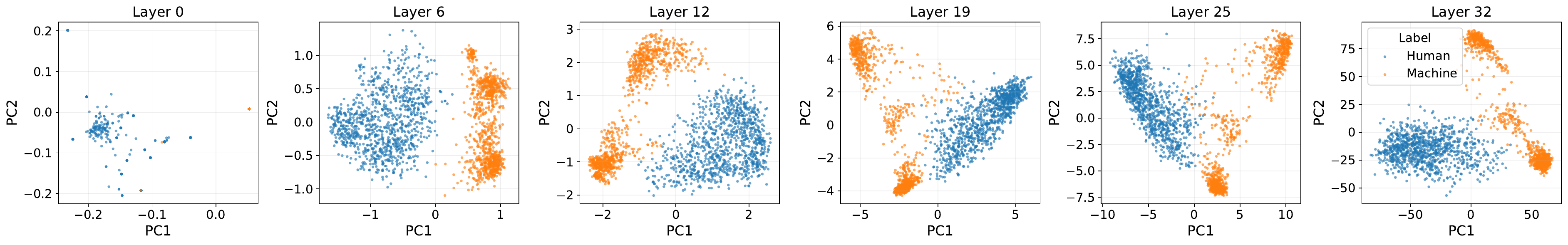}
        \caption{PeerRead}
        \label{fig:pca_peerread}
    \end{subfigure}
    
    \vspace{0.3cm}
    
    % ArXiv PCA
    \begin{subfigure}[b]{1\linewidth}
        \centering
        \includegraphics[width=\linewidth]{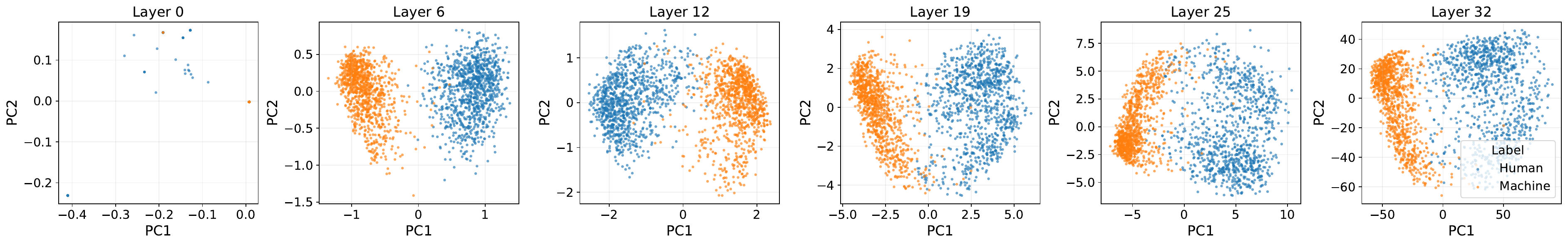}
        \caption{ArXiv}
        \label{fig:pca_arxiv}
    \end{subfigure}
    
    \caption{Projections of human- and machine-generated text hidden states onto the first two principal components across layers. This figure extends Figure~\ref{fig:layer_pca} to the Reddit, PeerRead, and ArXiv subsets of M4GT~\cite{wang2024m4gt}.}
    \label{app:fig:pca_domains}
\end{figure}

In Figure~\ref{fig:layer_pca}, we visualized the activations of human- and machine-written Wikipedia text projected onto the first two principal components.
Figure~\ref{app:fig:pca_domains} extends this analysis to the Reddit, PeerRead, and ArXiv subsets of M4GT~\cite{wang2024m4gt}.
Across all domains, we observe the same qualitative pattern: human- and machine-written texts occupy distinct regions of latent space and remain linearly separable throughout the middle and later layers.

\begin{figure}[h]
    \centering
    \begin{subfigure}[b]{0.49\linewidth}
        \centering
        \includegraphics[width=.8\linewidth]{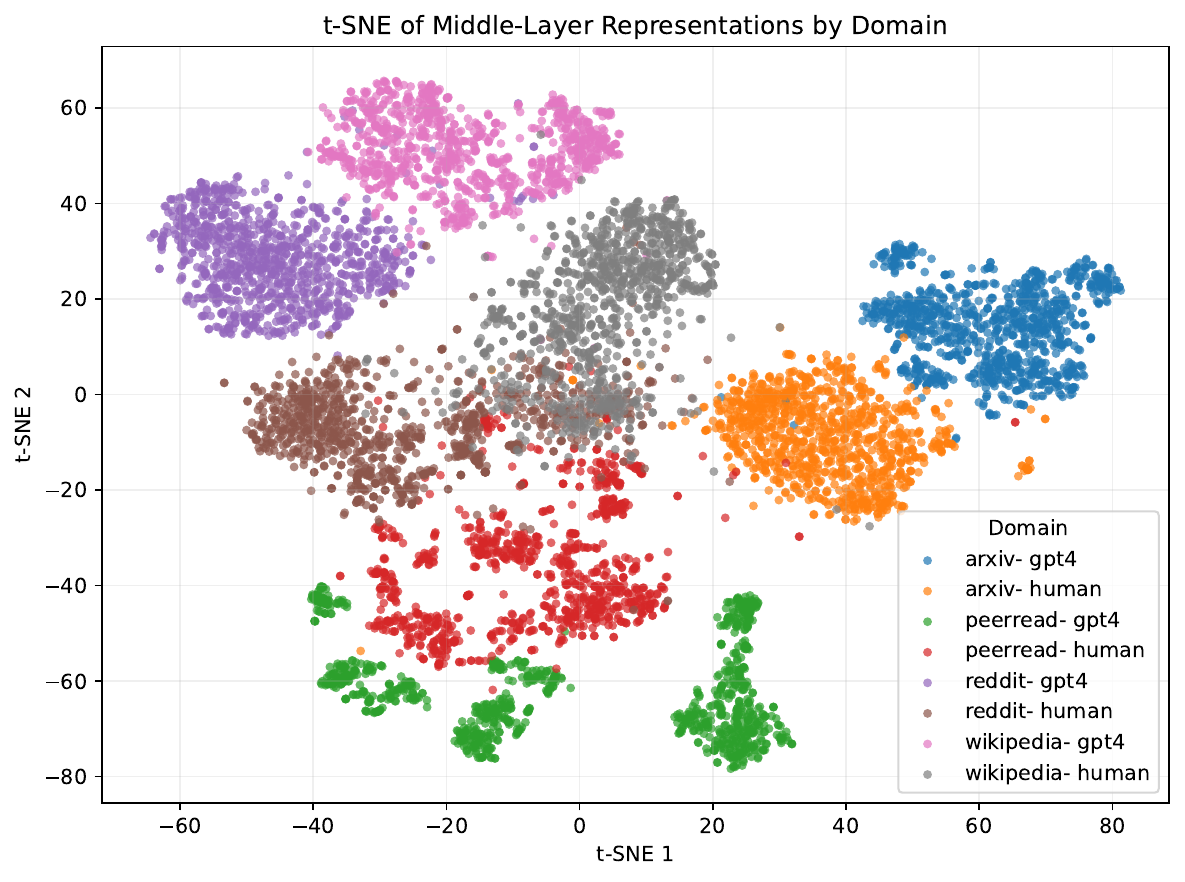}
        \caption{Domain comparison}
        \label{fig:map_domains}
    \end{subfigure}
    \hfill
    \begin{subfigure}[b]{0.49\linewidth}
        \centering
        \includegraphics[width=.8\linewidth]{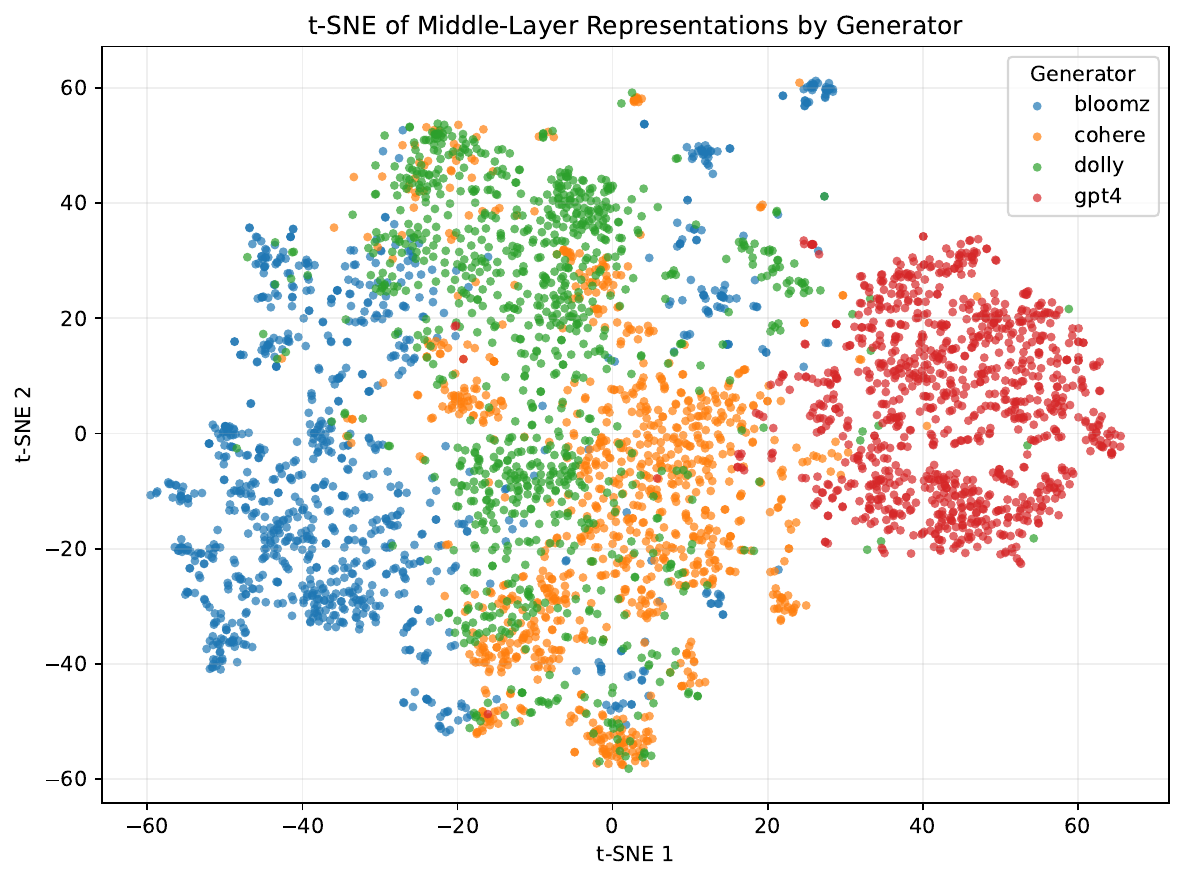}
        \caption{Generator comparison}
        \label{fig:map_generators}
    \end{subfigure}
    
    \caption{Low-dimensional t-SNE projections of human- and machine-written text from different domains (a) and generators (b) in the M4GT dataset~\cite{wang2024m4gt}.}
    \label{app:fig:map}
\end{figure}

In Figure~\ref{app:fig:map}, we visualize activations of human- and machine-written text from different domains (a) and generators (b) in M4GT~\cite{wang2024m4gt} using t-SNE.
In both plots, each domain or generator occupies a distinct local region of latent space, which further separates into human- and machine-written clusters.
Although t-SNE is nonlinear and hence does not preserve linear structure, the observed clustering is consistent with our earlier finding that human and machine representations are linearly separable within individual domains and generators.

\begin{figure}[h]
     \centering
     \includegraphics[width=\linewidth]{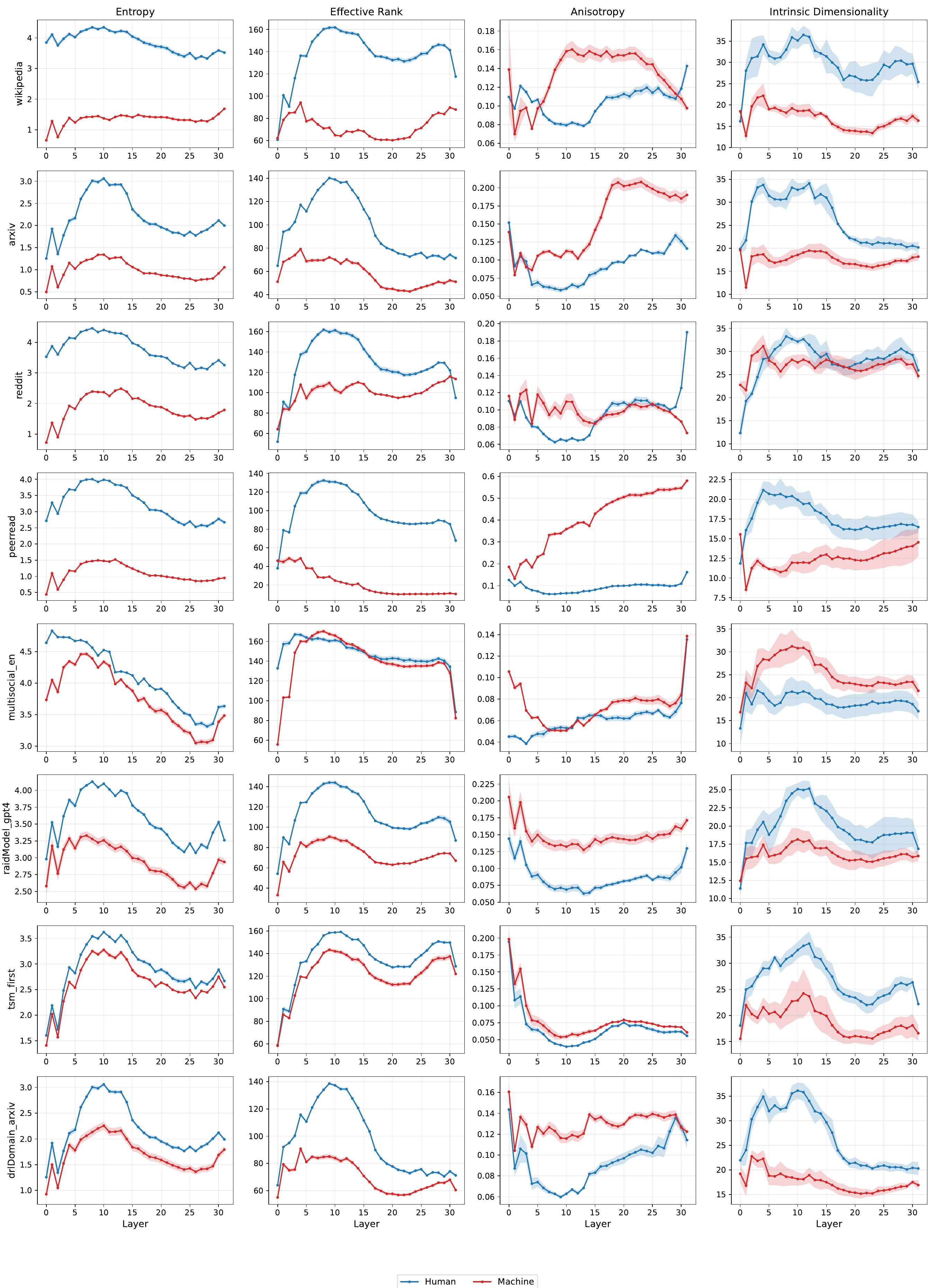}
    \caption{Representation-quality metrics for human and machine latent representations from the perspectives of information content (Entropy, Effective Rank) and geometric structure (Anisotropy, Intrinsic Dimensionality). Extended results to other domains than Wikipedia from M4GT~\cite{wang2024m4gt} and selected subsets from DetectRL~\cite{wu2024detectrl}, RAID~\cite{dugan2024raid}, MultiSocial~\cite{macko2025multi}, and TSM~\cite{quaremba2026tsm}.}
    \label{fig:f_qual}
\end{figure}

In Figure~\ref{fig:f_qual_wiki}, we analyzed the four representation quality metrics on the Wikipedia subset.
Figure~\ref{fig:f_qual} extends this analysis to the ArXiv, Reddit, and PeerRead subsets of M4GT~\cite{wang2024m4gt}, as well as selected subsets from DetectRL~\cite{wu2024detectrl}, RAID~\cite{dugan2024raid}, MultiSocial~\cite{macko2025multi}, and TSM~\cite{quaremba2026tsm}.
Across these diverse settings, we observe the same qualitative pattern as for Wikipedia: MGT representations exhibit lower entropy, effective rank, and intrinsic dimensionality, while showing higher anisotropy than HWT representations.
Although the differences are sometimes less pronounced, particularly in datasets that include adversarial attacks or multiple generators, the overall trend remains consistent.

%%%%%%%%%%%%%%%%%%%%%%%%%%%%%%%%%%%%%%%%%%%%%%%%%%%%%%%%%%%%%%%%%%%%%%%%
% EXPERIMENTAL SETUP
%%%%%%%%%%%%%%%%%%%%%%%%%%%%%%%%%%%%%%%%%%%%%%%%%%%%%%%%%%%%%%%%%%%%%%%%
\section{Experimental Setup}
\label{app:exp_setup}

\subsection{Benchmarks}

We briefly introduce each benchmark and describe how the data were sampled and split.
While we use M4GT~\cite{wang2024m4gt} for the representation analysis in Section~\ref{sec:mgt_rep}, initial experiments on its domain subsets revealed saturated results, with several baselines achieving near-perfect performance.
We therefore include more challenging benchmarks for the detection experiments.

\begin{itemize}
    \item \textbf{DetectRL}~\cite{wu2024detectrl}: improves upon prior benchmarks by constructing data that more closely reflects real-world detection scenarios, including a range of adversarial attacks such as prompt-based and paraphrase attacks.\footnote{\url{https://github.com/NLP2CT/DetectRL}} We use the Task 2 domain dataset and randomly sample training and test sets. Consequently, each domain split contains examples generated by different LLMs and subjected to diverse attack strategies.

    \item \textbf{MultiSocial}~\cite{macko2025multi}: focuses on social media text, which is typically shorter, noisier, and more linguistically diverse than text from traditional domains.
    The benchmark covers 22 languages across 5 social media platforms.\footnote{\url{https://github.com/kinit-sk/multisocial}}
    We randomly sample training and test sets from the full dataset, resulting in a diverse mixture of languages, text lengths, writing styles, and platforms.

    \item \textbf{RAID}~\cite{dugan2024raid}: is one of the largest MGT detection benchmarks, containing over 10 million text samples generated across diverse models, domains, decoding strategies, and adversarial attacks.\footnote{\url{https://github.com/liamdugan/raid}}
    We construct generator-specific subsets by randomly sampling from the full dataset. As a result, each subset contains a diverse mixture of domains, decoding strategies, and attack types.
    
    \item \textbf{TSM}~\cite{quaremba2026tsm}: focuses on the Wikipedia domain and text-generation tasks that closely resemble real-world editing behavior, including first-paragraph generation, paragraph extension, summarization, and text style transfer.\footnote{\url{https://github.com/gerritq/tsm_bench}}
    We use the English split and randomly sample training and test instances within each task across six generators.

\end{itemize}

For the experiments on detecting varying degress of AI-edited texts in Section~\ref{sec:exp:edit}, we use the following benchmarks:

\begin{itemize}
      \item \textbf{APT-Eval}~\cite{saha2025apt}: APT-Eval contains 15k AI-polished texts from diverse domains, covering varying degrees of AI involvement in human-written text.\footnote{\url{https://github.com/ShoumikSaha/ai-polished-text}}
    The benchmark evaluates five detectors and quantifies edit strength using semantic cosine similarity, Jaccard distance, and Levenshtein distance.
    We use these metrics to correlate MGT probe projection scores with the degree of AI editing.
    To obtain a balanced sample, we evenly draw 200 instances from each polishing category: (1) extremely minor, (2) minor, (3) slightly major, and (4) major.
    
    \item \textbf{EditLens}~\cite{thai2026editlens}: EditLens contains AI-edited human-written text from four domains: reviews, creative writing, general educational web articles, and news articles.\footnote{\url{https://github.com/pangramlabs/EditLens}}
    The benchmark uses three LLMs to edit human-written text with varying levels of intervention.
    We randomly sample from the full dataset using the same procedure as for APT-Eval.

\end{itemize}

\subsection{Baseline Detectors}

We test 16 MGT detectors covering zero-shot and supervised methods. 
Below we briefly introduce each detector and implementation details. If not otherwise stated, we implement each detector with Llama-3-8B~\cite{grattafiori2024llama3} for fair comparison.

\subsubsection*{Zero-shot}

\begin{itemize}

    \item \textbf{Likelihood}~\cite{solaiman2019loglikelihood}: A simple baseline that computes the average token log-likelihood of a text. Higher log-likelihood indicates that the text is more likely to have been generated by an LLM.

    \item \textbf{LLR}~\cite{su2023detectllm}: The Log-Likelihood Log-Rank Ratio (LLR) combines token log-likelihood and token rank. Log-likelihood captures a model's absolute confidence in the observed token, while rank captures its relative confidence. Higher LLR scores indicate that a text is more likely to be machine-generated.

    \item \textbf{Rank}~\cite{gehrmann2019gltr}: Assigns each token its rank under the language model's predicted token distribution. A lower average rank indicates that the observed tokens are more predictable and therefore more likely to have been generated by an LLM.

    \item \textbf{Revise}~\cite{zhu-etal-2023-beat}: A rewrite-based detector motivated by the observation that LLM-generated text changes less under rewriting than human-written text, as it already conforms closely to the statistical patterns learned by the model. Higher similarity between the original and rewritten text therefore indicates a higher likelihood of machine generation. Instead of ChatGPT, we use LLaMA-8B-Instruct\footnote{\url{https://huggingface.co/meta-llama/Meta-Llama-3-8B-Instruct}} for rewriting.

    \item \textbf{Binoculars}~\cite{hans2024binoculars}: Uses two LLMs to compute the ratio of perplexity to cross-perplexity. Cross-perplexity measures how surprising the token predictions of one model ("observer") are to another model ("performer"). Lower scores indicate that a text is more likely to be machine-generated.
    
    \item \textbf{FastDetectGPT}~\cite{bao2023fast}: An efficient variant of DetectGPT~\cite{mitchell2023detectgpt}. DetectGPT exploits the observation that LLM-generated text tends to lie in regions of negative curvature of the model's log-probability surface. FastDetectGPT approximates this signal more efficiently through sampling-based estimation.
    
    \item \textbf{GECScore}~\cite{wu2025wrote}: Computes the Grammar Error Correction Score (GECScore), motivated by the observation that human-written text typically contains more grammatical errors than LLM-generated text. Following the original method, the score is derived from the similarity between the original text and its grammar-corrected version. Due to computational constraints, we use the 8B model variant instead of LLaMA-3-70B.

    \item \textbf{RAIDAR}~\cite{mao2024raidar}: Another rewrite-based detector motivated by the observation that LLMs tend to modify human-written text more substantially than machine-generated text.
    The method derives an MGT signal by measuring the edit distance between the original text and its rewritten version.
    
\end{itemize}

\subsubsection*{Supervised detector}

\begin{itemize}
    
    \item \textbf{OpenAI-RoBERTa}~\cite{solaiman2019loglikelihood}: A RoBERTa~\cite{liu2019roberta} classifier fine-tuned to distinguish GPT-2-generated text\footnote{\url{https://huggingface.co/openai-community/gpt2-xl}} from human-written WebText data.

    \item \textbf{ID}~\cite{hu2023radar}: Uses the Persistent Homology Dimension (PHD) estimator~\cite{schweinhart2021persistent} to measure the intrinsic dimensionality of text representations. The resulting scalar feature is fed into a logistic regression classifier. We use XLM-RoBERTa~\cite{conneau2020} as the backbone model.
    
    \item \textbf{RADAR}~\cite{hu2023radar}: Trains a detector through adversarial learning between a paraphraser and a discriminator, using Vicuna-7B~\cite{zheng2023vicuna} as the base model.

    \item \textbf{BiScope}~\cite{guo2024biscope}: Motivated by the observation that LLMs tend to memorize local context differently for human- and machine-generated text. The method computes cross-entropy losses between output logits and both the ground-truth token and the immediately preceding token. A classifier is then trained on statistics derived from these losses.

    \item \textbf{TextFluoroscopy}~\cite{yu2024fluo}: Argues that the strongest MGT signals emerge in middle layers. The method identifies the middle-layer representation whose vocabulary-space distribution differs most from the first and last layers and trains a nonlinear MLP classifier on that representation. We replace the originally used GTE-Qwen1.5-7B-Instruct model\footnote{\url{https://huggingface.co/Alibaba-NLP/gte-Qwen1.5-7B-instruct}} with LLaMA-3-8B.

    \item \textbf{RepreGuard}~\cite{chen2025repreguard}: Identifies a probing direction by computing the first principal component of the activation differences between human- and machine-generated text. Although RepreGuard is training-free, it still requires labeled examples to estimate the probing direction through PCA.

    \item \textbf{EditLens}~\cite{thai2026editlens}: Trains a regression model to estimate the degree of AI editing in a text. We use the released RoBERTa-Large model,\footnote{\url{https://huggingface.co/pangram/editlens_roberta-large}} which was fine-tuned on data spanning different levels of AI-edit strength.

    \item \textbf{RoBERTa}~\cite{liu2019roberta}: We fully fine-tune RoBERTa-Base\footnote{\url{https://huggingface.co/FacebookAI/roberta-base}} on DetectRL, TSM, and RAID. For MultiSocial, we instead fine-tune XLM-RoBERTa-Base\footnote{\url{https://huggingface.co/FacebookAI/xlm-roberta-base}} to support multilingual inputs. We use a batch size of 32, a learning rate of $2\times10^{-5}$, weight decay of 0.01, and train for two epochs.
\end{itemize}
%%%%%%%%%%%%%%%%%%%%%%%%%%%%%%%%%%%%%%%%%%%%%%%%%%%%%%%%%%%%%%%%%%%%%%%%
% Related work
%%%%%%%%%%%%%%%%%%%%%%%%%%%%%%%%%%%%%%%%%%%%%%%%%%%%%%%%%%%%%%%%%%%%%%%%
\subsection{Comparison to Related Work Detectors}
\label{app:exp_setup:comp}

\paragraph{Linear Separability and Representation Quality Analyses}
Compared to prior work on MGT detection based on latent representations~\cite{tulchinskii2023intrinsic,yu2024fluo,chen2025repreguard}, our study differs both in its analysis of latent representations and in its methodological approach.
\citet{tulchinskii2023intrinsic} were among the first to investigate latent-space differences between human- and machine-written text.
While they show that intrinsic dimensionality differs between the two, we additionally analyze entropy, effective rank, and anisotropy, examine their evolution across layers, and demonstrate that all four metrics exhibit consistent patterns across diverse domains.
TextFluoroscopy~\cite{yu2024fluo} provides evidence that middle layers contain particularly robust MGT signals, but does not investigate the linear separability of human and machine representations.
RepreGuard~\cite{chen2025repreguard} identifies systematic differences in activation patterns, but neither tests for linear separability nor connects its findings to the Linear Representation Hypothesis (LRH)~\cite{park2023linear}.

\paragraph{Methodology}
Methodologically, our approach differs in several ways.
ID~\cite{tulchinskii2023intrinsic} trains a logistic regression classifier on a single intrinsic-dimensionality feature per sample, whereas our probes operate directly on latent representations.
TextFluoroscopy~\cite{yu2024fluo} trains a \textit{nonlinear} classifier on the middle-layer representation selected through vocabulary-space distributional differences between early and late layers.
In contrast, we train simple \textit{linear} classifiers and show that they achieve stronger performance.
RepreGuard~\cite{chen2025repreguard} identifies probing directions through an unsupervised, \textit{training-free} procedure.
Specifically, it computes the first principal component of the activation-difference matrix between human and machine text using mean-pooled token representations.\footnote{This is closely related and often identical to difference-in-means~\cite{hollinsworth2024language,rimsky2024steering}.}
In contrast, we \textit{train} regularized linear classifiers directly on representations and show that last-token pooling consistently outperforms mean pooling (Appendix~\ref{app:ablation}).
Moreover, RepreGuard operates in the full activation space (e.g., 4,096 dimensions for LLaMA-8B), whereas we first project activations into a 100-dimensional PCA subspace.
As our representation analysis shows that MGT and HWT remain separable in low-dimensional spaces, our probes operate on only 2.4\% of the original dimensionality, substantially reducing memory and computational requirements.

Tables~\ref{tab:t_id} and~\ref{tab:t_ood} show that MGT probes consistently outperform these related methods.
Compared to RepreGuard~\cite{chen2025repreguard}, we attribute these gains primarily to supervised, regularized learning, which suppresses noisy dimensions and yields more robust probing directions.
Compared to TextFluoroscopy~\cite{yu2024fluo}, our probes avoid unnecessary nonlinear complexity and leverage information from all layers rather than relying on a single middle layer.
As shown in Section~\ref{sec:mgt_rep}, the boundary between human- and machine-written text is largely linear, while Appendix Figure~\ref{fig:f_layer} demonstrates that aggregating information across layers improves robustness.
Finally, unlike ID~\cite{tulchinskii2023intrinsic}, our probes operate directly on latent representations rather than a single-metric feature, enabling them to exploit substantially richer information about machine-generated text.

\subsection{Computational Resources}

We run all experiments on an HPC cluster equipped with NVIDIA A100 (40GB) and H200 (141GB) GPUs.
Training a single probe variant, including activation extraction and inference, typically requires only 1--3 minutes on an A100 (40GB).

\section{Additional Results}
\label{app:results}

\subsection{ID Adversarial Attacks}

\begin{table}[h]
    \centering
    {\renewcommand{\arraystretch}{1.0}
    \begin{adjustbox}{width=\linewidth}
        \begin{tabular}{lcccc}
\toprule
& \multicolumn{4}{c}{\textbf{DetectRL Attacks~\citep{wu2024detectrl}}} \\
\cmidrule(lr){2-5}
\textbf{Model} & \textbf{Mixing} & \textbf{Paraphrase} & \textbf{Perturbation} & \textbf{Prompt} \\
\midrule
\multicolumn{5}{l}{\textbf{Zero-shot}} \\
\midrule
Likelihood & 0.9562 & 0.6973 & 0.6535 & 0.9628 \\
LLR & 0.9232 & 0.6813 & 0.6876 & 0.9288 \\
Rank & 0.2121 & 0.4113 & 0.8194 & 0.1584 \\
Binoculars & 0.9992 & 0.9069 & 0.9248 & 0.9847 \\
FastDetectGPT & 0.8911 & 0.7858 & 0.6218 & 0.8926 \\
GECScore & 0.7183 & 0.6777 & 0.6075 & 0.7064 \\
RAIDAR & 0.7392 & 0.7209 & 0.6503 & 0.7122 \\
\midrule
\multicolumn{5}{l}{\textbf{Supervised}} \\
\midrule
OpenAI-RoBERTa & 0.8774 & 0.8991 & 0.6946 & 0.8581 \\
RADAR & 0.9733 & 0.9628 & 0.9593 & 0.9474 \\
EditLens & 0.3302 & 0.5628 & 0.5775 & 0.3562 \\
ID & 0.7089 & 0.7253 & 0.6406 & 0.6746 \\
RepreGuard & 0.9600 & 0.8480 & 0.9820 & 0.9260 \\
BiScope & \cellcolor{orange!25}\underline{0.9993} & 0.9692 & 0.9884 & 0.9946 \\
TextFluoroscopy & 0.9944 & 0.9968 & \cellcolor{orange!25}\underline{0.9971} & 0.9934 \\
RoBERTa & 0.9964 & 0.9992 & \cellcolor{cyan!25}\textbf{1.0000} & 0.9972 \\
\midrule
\multicolumn{5}{l}{\textbf{MGT Probes}} \\
\midrule
LLP & \cellcolor{orange!25}\underline{0.9993} & \cellcolor{cyan!25}\textbf{0.9999} & \cellcolor{cyan!25}\textbf{1.0000} & \cellcolor{orange!25}\underline{0.9982} \\
\hspace*{1em}$\Delta$ vs BL & \textcolor{green!60!black}{+0.00} & \textcolor{green!60!black}{+0.07} & \textcolor{green!60!black}{+0.00} & \textcolor{green!60!black}{+0.10} \\
CLP & \cellcolor{cyan!25}\textbf{0.9995} & \cellcolor{orange!25}\underline{0.9998} & \cellcolor{cyan!25}\textbf{1.0000} & \cellcolor{cyan!25}\textbf{0.9994} \\
\hspace*{1em}$\Delta$ vs BL & \textcolor{green!60!black}{+0.02} & \textcolor{green!60!black}{+0.06} & \textcolor{green!60!black}{+0.00} & \textcolor{green!60!black}{+0.22} \\
\bottomrule
\end{tabular}

    \end{adjustbox}}
    \caption{In-domain detection AUC scores for adversarial attacks on DetectRL~\cite{wu2024detectrl}. LLP = Layer-Averaged Linear Probes. CLP=Concatenated-Layer Probes. For TSM, FP=First Paragraph, PE=Paragraph Extension, SUM=Summarization, and TST=Text Style Transfer. $\Delta$ denotes the gain over the strongest baseline.}
    \label{tab:t_id_attacks}
\end{table}

In addition to the dimensions considered in Table~\ref{tab:t_id} (domains, languages, generators, and generation tasks), we also evaluate adversarial attacks using the DetectRL benchmark~\cite{wu2024detectrl}.
We observe the same saturation effects as for the DetectRL domain subsets, with most baselines achieving near-perfect performance.
For this reason, we omit adversarial attacks from the main table.

\subsection{OOD}
\label{app:ood}

\begin{figure}[h]
     \centering
     \includegraphics[width=\linewidth]{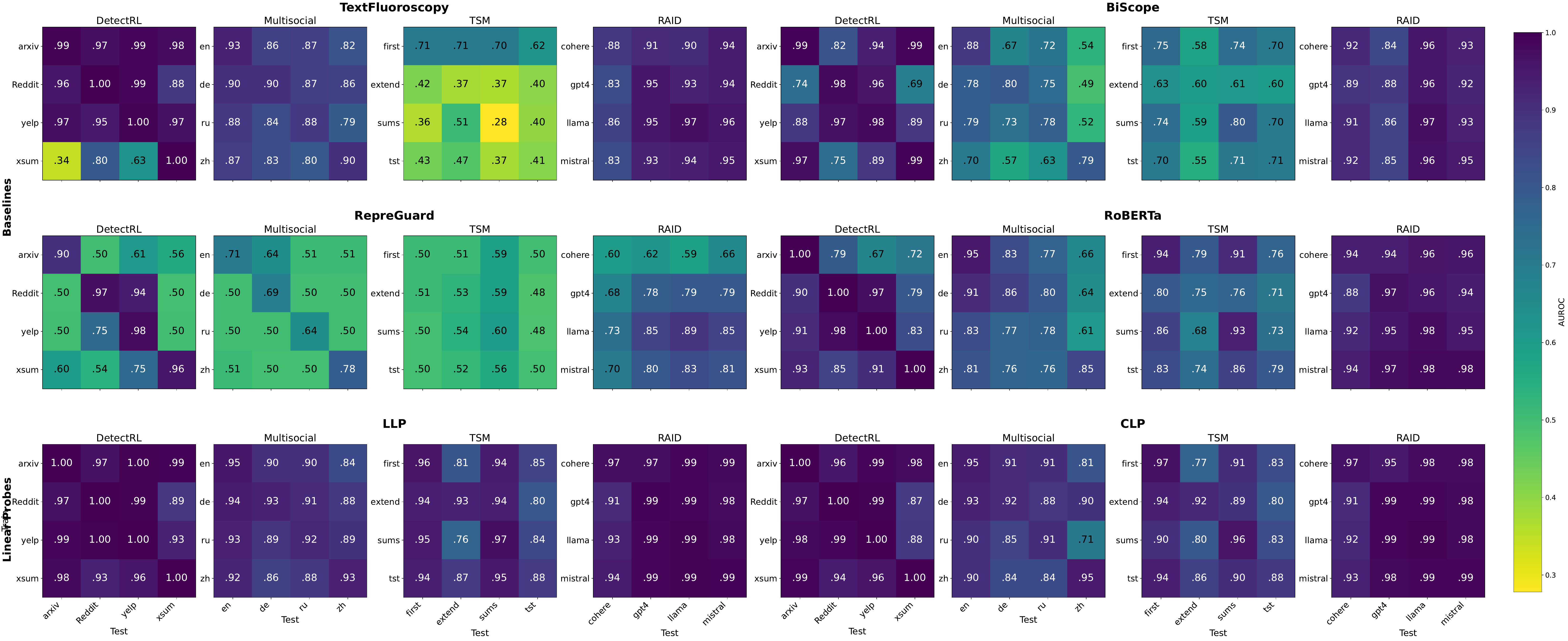}
    \caption{OOD AUC scores for baseline detectors (top rows) and MGT probes (bottom row). Rows denote the training subsets, while columns denote the test subsets. Darker colors indicate better performance.}
    \label{fig:f_ood}
\end{figure}

In Table~\ref{tab:t_ood}, we report the mean AUC for each target subset unseen during training, averaged over transfers from the remaining subsets within the same benchmark. Figure~\ref{fig:f_ood} provides the full OOD transfer results.

\subsection{Probing vector similarity}
\label{app:sim}

\begin{figure}[h]
     \centering
     \includegraphics[width=1\linewidth]{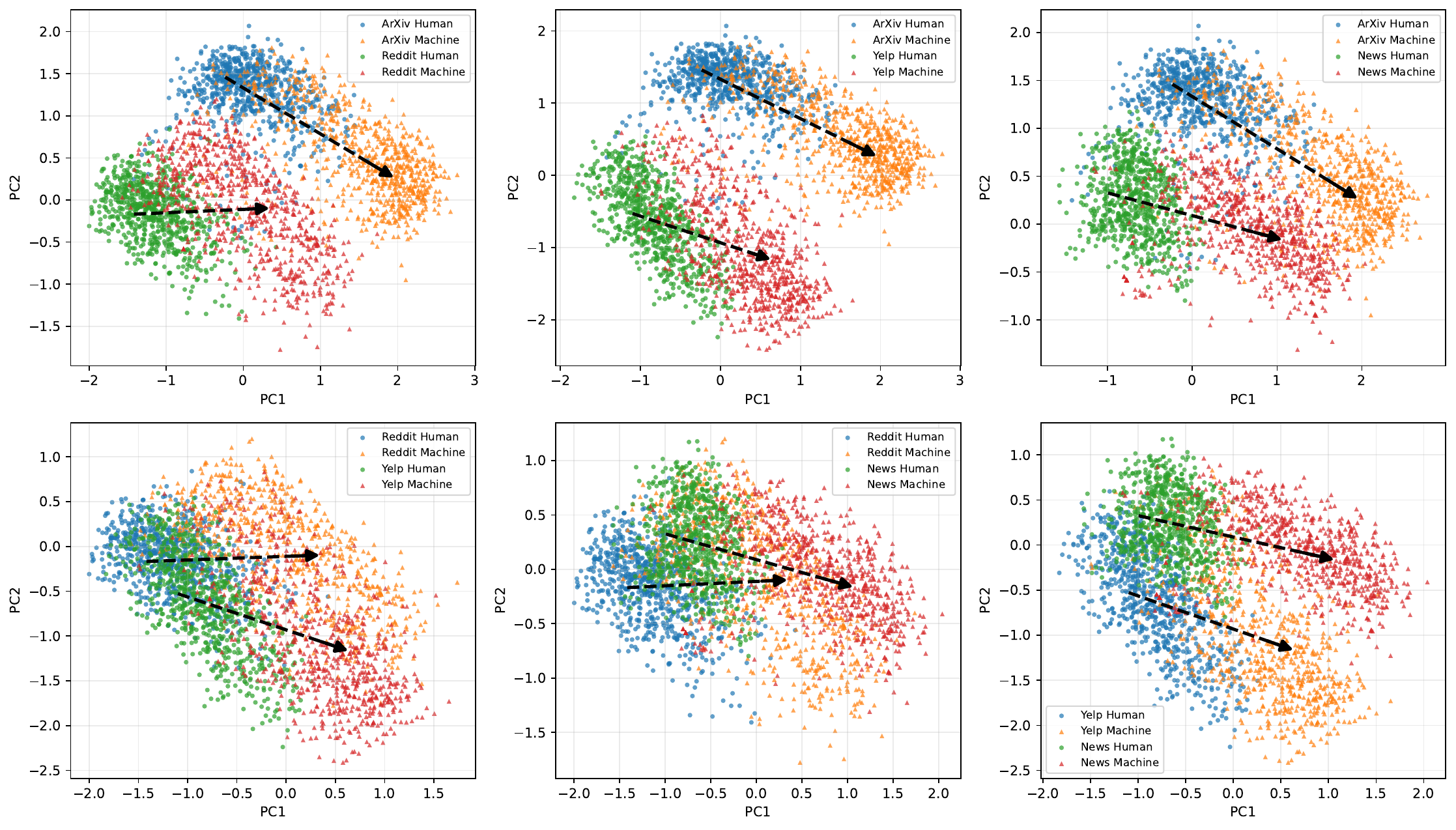}
    \caption{Activation projections of paired DetectRL datasets~\cite{wu2024detectrl} in the shared PCA space at layer 16. Dashed lines connect the centroids of the human and machine clusters within each dataset.}
    \label{fig:f_drl_parallel}
\end{figure}

Figure~\ref{fig:f_drl_parallel} shows layer-16 activation projections for paired domain subsets in DetectRL~\cite{wu2024detectrl}.
The black dashed line connects the centroids of the human and machine activation clouds within each subset, approximating the corresponding linear MGT direction.

Consistent with Figure~\ref{fig:f_probes_pca}, we find that these linear MGT directions are broadly aligned in the same direction across most subsets.
They are not perfectly aligned, which is expected because (\textit{1}) we approximate the MGT direction by connecting class centroids rather than learning it directly, and (\textit{2}) the subsets differ along several confounding dimensions, including adversarial attacks, text length, and stylistic or grammatical variation.
The latter is particularly evident for Reddit, which exhibits the largest distributional shift across domains.
Nevertheless, their broad alignment provides further evidence that probes learn an approximate shared linear direction associated with MGT.

\subsection{Sample-efficiency Analysis}
\label{app:sample_eff}

\begin{figure}[h]
     \centering
     \includegraphics[width=.8\linewidth]{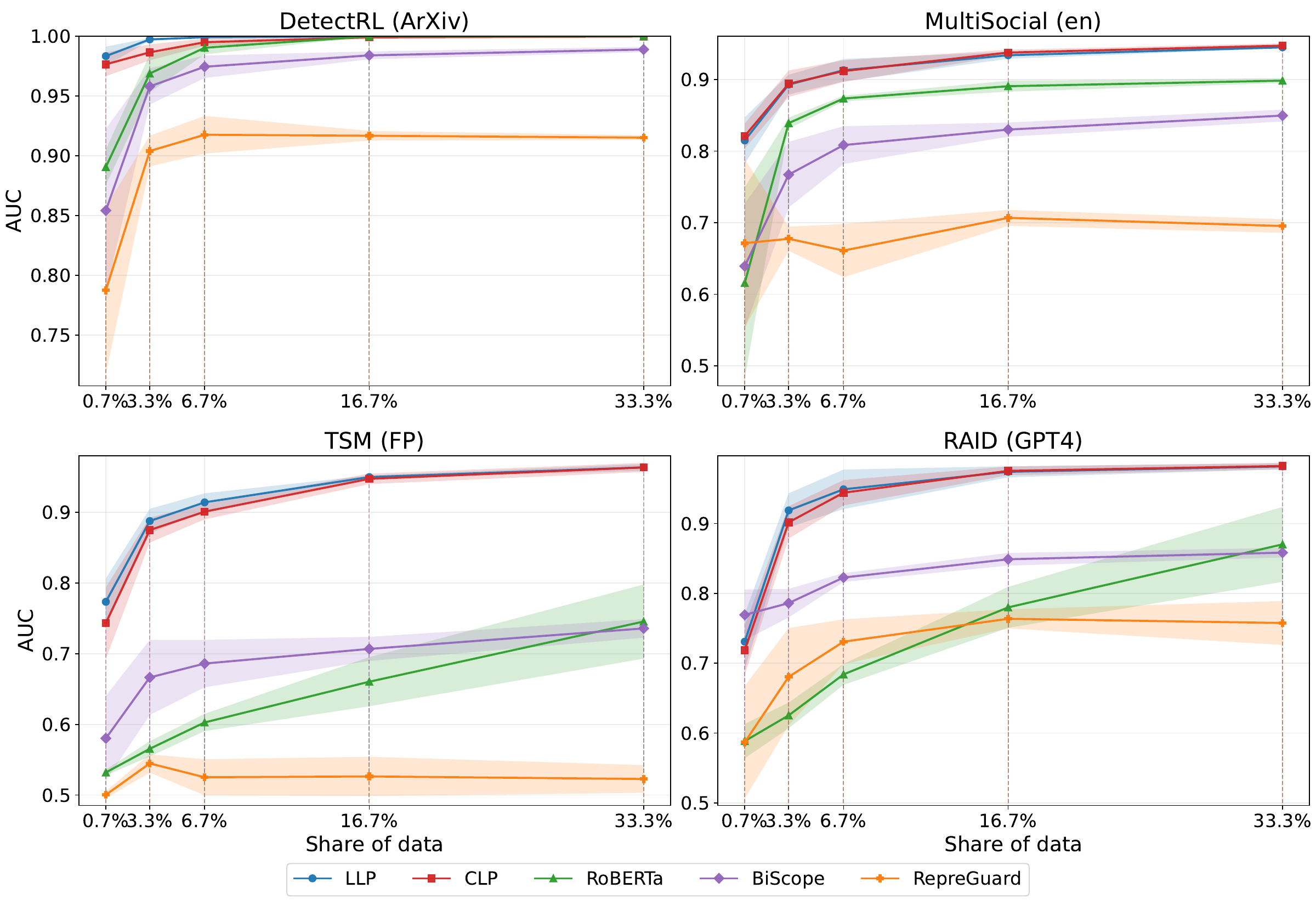}
    \caption{In-domain AUC scores for selected datasets as a function of training set size. Compared to Figure~\ref{fig:f_samples_pca}, we do not reduce the dimensionality of activations.}
    \label{app:fig:f_samples}
\end{figure}

In Figure~\ref{fig:f_samples_pca}, we evaluate sample efficiency using a PCA space constructed from the full training set for each layer, and then train probes on progressively smaller subsets.
However, this setup does not fully reflect a real-world application, where the full training distribution may not be available in advance.
Therefore, in Figure~\ref{app:fig:f_samples}, we repeat the experiment in the full activation space, without applying PCA.
We observe the same pattern, confirming that the sample-efficiency results.

\subsection{Detecting AI-edited text}
\label{app:edits}

\begin{figure}[h]
     \centering
     \includegraphics[width=\linewidth]{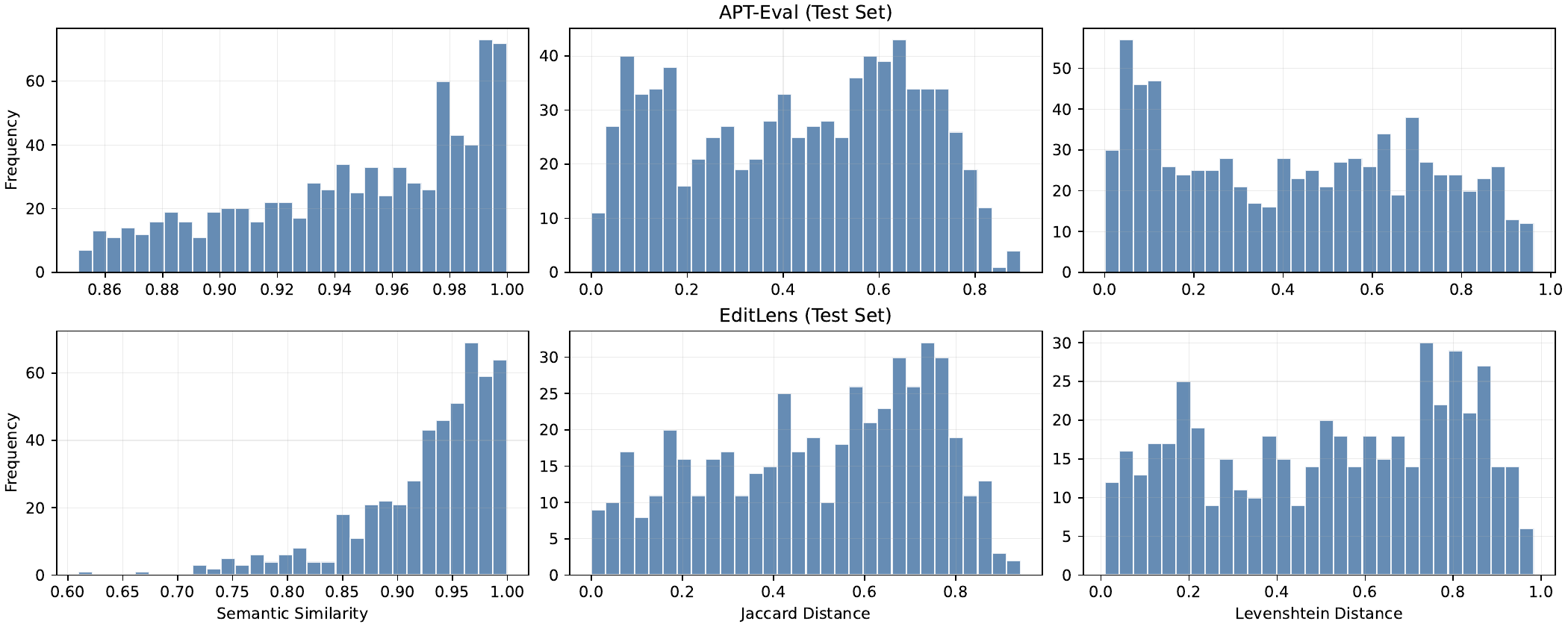}
    \caption{Histograms of the similarity measures Semantic (Cosine) Similarity, Jaccard Distance, and Levenshtein Distance for the test sets of APT~\cite{saha2025apt} and EditLens~\cite{thai2026editlens}.}
    \label{fig:hist_edits}
\end{figure}

In Section~\ref{sec:exp:edit}, we investigate whether probing vectors encode MGT signals as a continuum rather than a binary distinction.
Figure~\ref{fig:hist_edits} shows the distributions of the text-similarity metrics used to quantify the degree of AI editing.
Our random sampling procedure yields a relatively uniform distribution across metrics and benchmarks, with the exception of semantic similarity in EditLens, which exhibits a left-skewed distribution.

\begin{figure}[h]
     \centering
     \begin{subfigure}[b]{1\linewidth}
         \centering
         \includegraphics[width=\linewidth]{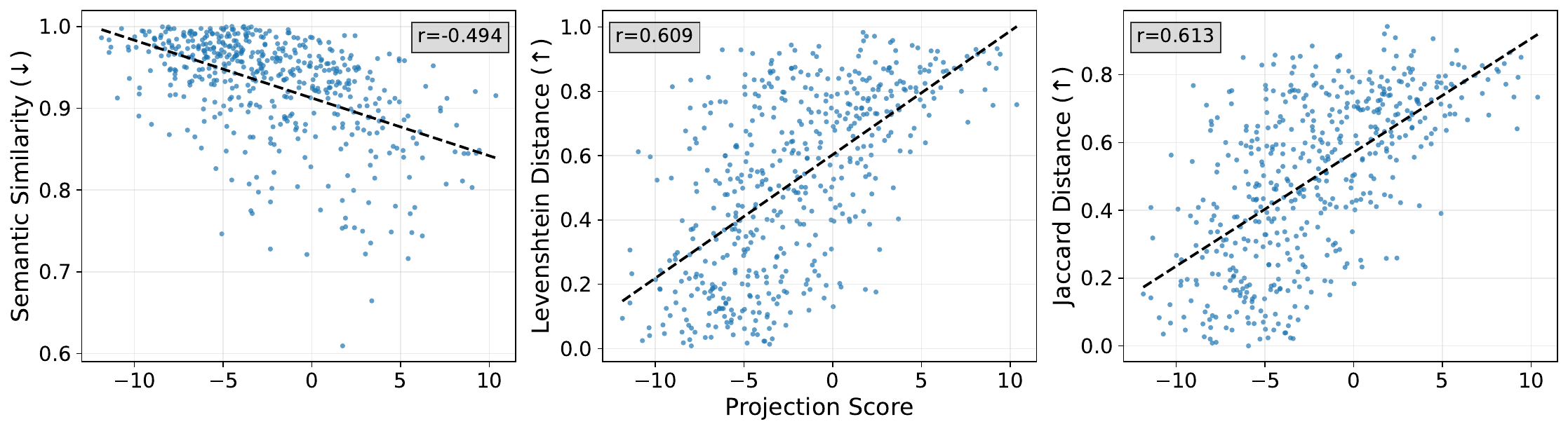}
         \caption{LLP}
     \end{subfigure}
     
      \vfill \vspace{0.3cm}
     
     \begin{subfigure}[b]{1\linewidth}
         \centering
         \includegraphics[width=\linewidth]{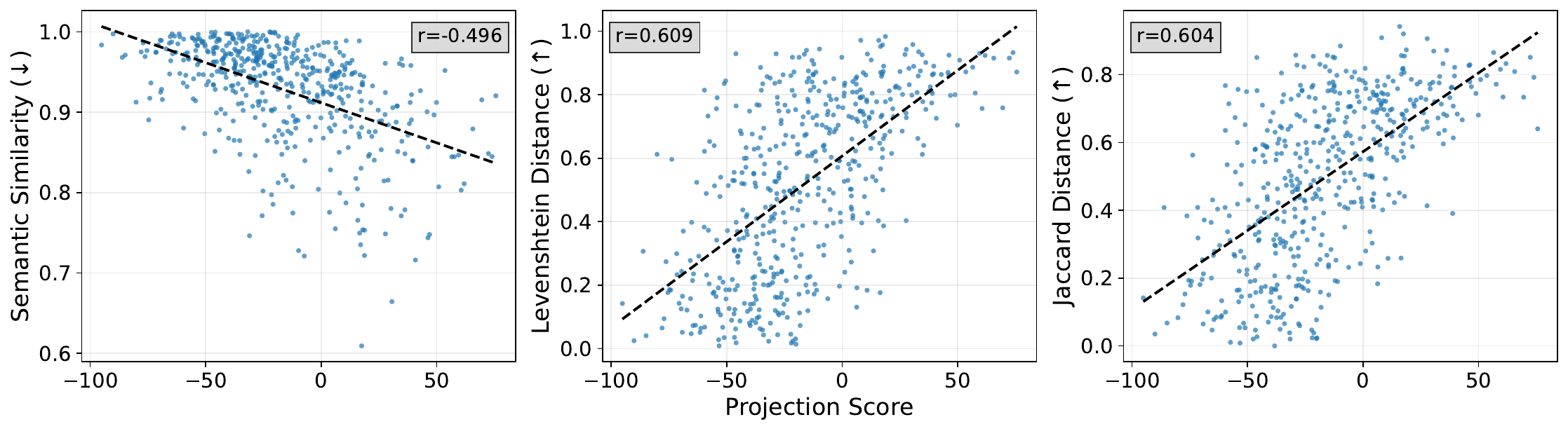}
         \caption{CLP}
     \end{subfigure} 
    \caption{Edit-strength metrics plotted against probe projection scores. The dashed line shows the linear regression fit, and $r$ denotes the Pearson correlation coefficient. This Figure shows the results for EditLens~\cite{thai2026editlens}.}
    \label{fig:f_edit_editlens}
\end{figure}

In Figure~\ref{fig:f_edit_apt}, we plot the text-similarity metrics used to quantify the degree of AI editing in APT-Eval~\cite{saha2025apt} against the projection scores of LLP and CLP.
Figure~\ref{fig:f_edit_editlens} presents the same analysis for EditLens~\cite{thai2026editlens}.
We observe the same overall pattern across both benchmarks, although the performance gap between LLP and CLP is smaller on EditLens.

%%%%%%%%%%%%%%%%%%%%%%%%%%%%%%%%%%%%%%%%%%%%%%%%%%%%%%%%%%%%%%%%%%%%%%%%
% LAYER ANALYSIS
%%%%%%%%%%%%%%%%%%%%%%%%%%%%%%%%%%%%%%%%%%%%%%%%%%%%%%%%%%%%%%%%%%%%%%%%
\subsection{Layer Analysis}
\label{sec:exp:layer}

\begin{figure}[h] 
     \centering
     % Left subfigure
     \begin{subfigure}[b]{0.48\linewidth}
         \centering
         \includegraphics[width=\linewidth]{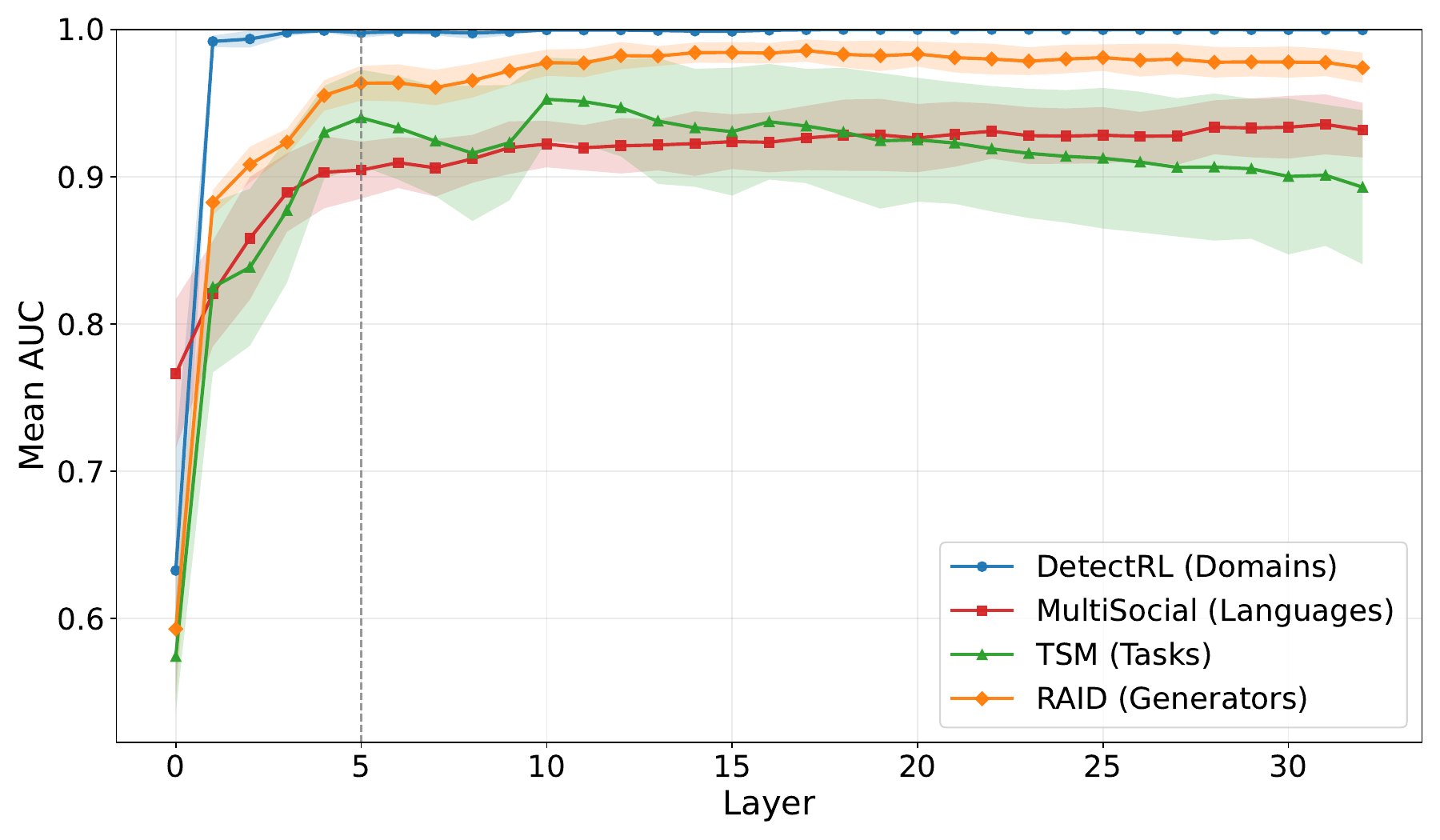}
         \caption{In-domain}
         \label{fig:layer_id}
     \end{subfigure}
     \hfill % Adds flexible space between the two
     % Right subfigure
     \begin{subfigure}[b]{0.48\linewidth}
         \centering
         \includegraphics[width=\linewidth]{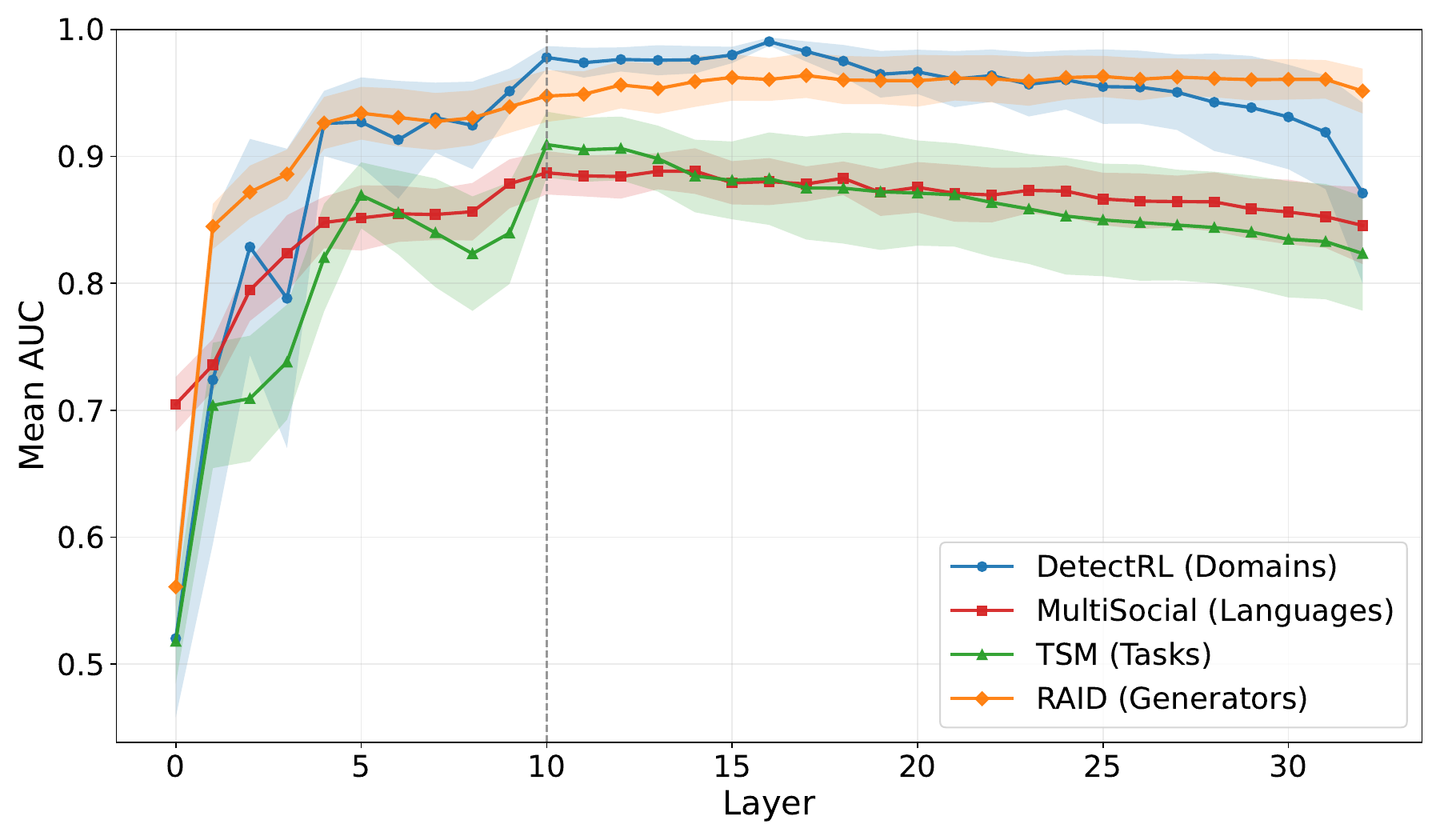}
         \caption{Out-of-domain}
         \label{fig:layer_ood}
     \end{subfigure}
     
    \caption{Layer-wise AUC averaged across benchmarks. Vertical lines denote performance plateaus. \textbf{Takeaway}: MGT signals emerge early and remain stable.} 
    \label{fig:f_layer}
\end{figure}

% \begin{figure}[t] % 't' (top) is preferred over 'h'/'H' for ACL/EMNLP formatting
%      \centering
%      \begin{subfigure}[b]{1\linewidth}
%          \centering
%          \includegraphics[width=\linewidth]{items/f_layer_id_default.pdf}
%      \end{subfigure}
     
%       \vfill \vspace{0.3cm}
     
%      \begin{subfigure}[b]{1\linewidth}
%          \centering
%          \includegraphics[width=\linewidth]{items/f_layer_ood_default.pdf}
%      \end{subfigure} 
%     \caption{Layer-wise AUC averaged across benchmark subsets for in-domain (top) and OOD (bottom) settings. Vertical lines denote performance plateaus. \textbf{Takeaway}: MGT signals emerge relatively early (around layer 5/8) and remain stable throughout the remaining layers.}
%     \label{fig:f_layer}
% \end{figure}

We conduct an additional analysis of probe performance across transformer layers to better understand where MGT signals emerge and how they evolve throughout the network.
Specifically, we ask: \textbf{RQ}: \textit{How does MGT detection performance evolve across layers in in-domain and OOD settings?}

\paragraph{MGT signals emerge in early layers and remain relatively stable throughout later layers.}
Figure~\ref{fig:f_layer} shows layer-wise mean AUC across the four subsets within each benchmark for both ID and OOD settings.
Both settings reveal a consistent two-stage pattern across benchmarks: a sharp rise in MGT signals in the early layers (up to approximately layer 5 for ID and layer 10 for OOD), followed by a plateau in which detection performance remains relatively stable.
In both settings, this suggests that even early representations associated with lower-level textual features~\cite{tenney2019bert} encode strong linear MGT signals.
For OOD detection, the strong performance of middle layers and the degradation in the final layers are consistent with later representations becoming increasingly specialized and therefore less generalizable~\cite{skean2025layer}.

%%%%%%%%%%%%%%%%%%%%%
% Disucssion
%%%%%%%%%%%%%%%%%%%%%
% How to these layer-wise performance findings compare to prior LP?
% Cite 1-2 papers
% Layer optimization paper: find high performance from layer 8 onwards
% Layer by layer paper also shows middle are the best!
% text fluo does this too!

%%%%%%%%%%%%%%%%%%%%%%%%%%%%%%%%%%%%%%%%%%%%%%%%%%%%%%%%%%%%%%%%%%%%%%%%
% TEXT LENGTH
%%%%%%%%%%%%%%%%%%%%%%%%%%%%%%%%%%%%%%%%%%%%%%%%%%%%%%%%%%%%%%%%%%%%%%%%
\subsection{Sensitivity to Text Length}
\label{sec:exp:length}

\begin{table}[h]
    \centering
    {\renewcommand{\arraystretch}{1.0}
    \begin{adjustbox}{width=1\linewidth}
        \begin{tabular}{lccccccccccc}
\toprule
Model $\downarrow$ & \multicolumn{11}{c}{$n$ chars $\rightarrow$} \\
 & 25 & 50 & 75 & 100 & 125 & 150 & 175 & 200 & 250 & 300 & 400 \\
\midrule
LLP & 0.850 & 0.823 & 0.881 & 0.908 & 0.966 & 0.981 & 0.989 & 0.993 & 0.991 & 0.995 & 0.997 \\
CLP & 0.840 & 0.824 & 0.903 & 0.960 & 0.982 & 0.990 & 0.996 & 0.996 & 0.997 & 0.997 & 1.000 \\
RoBERTa & 0.874 & 0.883 & 0.904 & 0.969 & 0.926 & 0.936 & 0.998 & 0.999 & 1.000 & 0.999 & 1.000 \\
\bottomrule
\end{tabular}

    \end{adjustbox}}
    \caption{Ablation study of text length. We compare LLP and CLP against a RoBERTa baseline. Each model is trained on the number of characters indicated by the corresponding column, using the ArXiv subset of DetectRL~\cite{wu2024detectrl}.}
    \label{tab:t_length}
\end{table}

Table~\ref{tab:t_length} presents AUC detection performance as a function of text length. We compare both MGT probes with the RoBERTa~\cite{liu2019roberta} baseline, training each model on the ArXiv subset of DetectRL~\cite{wu2024detectrl}. While RoBERTa has a slight AUC advantage for texts up to 50 characters, both probes remain competitive and reach near-peak performance at around 125 characters. These results indicate that the probes remain reliable even for relatively short texts.

%%%%%%%%
% AI-EDITING BENCHMARK
%%%%%%%%%%%%%%%%%%%%%%%%%%%%%%%%%%%%%%%%%%%%%%%%%%%%%%%%%%%%%%%%%%%%%%%%
\subsection{MGT Probes performance on AI-edit text detection}
\label{sec:exp:ai_edits}

\begin{table}[h]
    \centering
    {\renewcommand{\arraystretch}{1.0}
    \begin{adjustbox}{width=.9\linewidth}
        \begin{tabular}{lccccc}
\toprule
Model & Accuracy & Macro F1 & Human F1 & AI-edited F1 & AI-generated F1 \\
\midrule
RoBERTa & 0.652 & 0.642 & 0.583 & 0.473 & 0.869 \\
EditLens & 0.525 & 0.438 & 0.575 & 0.739 & 0.000 \\
LLP & 0.848 & 0.848 & 0.806 & 0.754 & 0.985 \\
CLP & 0.854 & 0.853 & 0.811 & 0.764 & 0.985 \\
\bottomrule
\end{tabular}

    \end{adjustbox}}
    \caption{Three-way classification on EditLens~\cite{thai2026editlens}, comparing the EditLens Llama-3.2-3B model, RoBERTa~\cite{liu2019roberta}, and both MGT probes. The three classes are ``AI-generated'', ``AI-edited'', and ``Human''.}
    \label{tab:t_ai_edit}
\end{table}

Figure~\ref{fig:f_edit_apt} shows a positive correlation between edit strength and the probe projection scores. In Table~\ref{tab:t_ai_edit}, we present initial results on three-way AI-edit classification using EditLens~\cite{thai2026editlens}. We compare two baselines, RoBERTa~\cite{liu2019roberta} and the EditLens Llama-3.2-3B model,~\footnote{\url{https://huggingface.co/pangram/editlens_Llama-3.2-3B}} against our MGT probes.

We split the EditLens data into train/validation/test sets of 1,000/200/200 examples. We train the probes and RoBERTa on the training set and calibrate all three methods on the validation set by selecting two thresholds for the classes ``AI-generated'', ``AI-edited'', and ``Human''. We do not further train the EditLens model, as it was already trained on this dataset.

Table~\ref{tab:t_ai_edit} reports accuracy and class-wise F1 scores. Both probes achieve strong classification performance relative to the baselines, despite not being explicitly trained for three-way classification. While these results indicate that the probe scores may support fine-grained AI-edit classification, the analysis is preliminary and does not support broader conclusions. We leave a more comprehensive evaluation of this direction to future work.

%%%%%%%%%%%%%%%%%%%%%%%%%%%%%%%%%%%%%%%%%%%%%%%%%%%%%%%%%%%%%%%%
%%%%%%%%
% CALIBRATION ANALYSIS
%%%%%%%%%%%%%%%%%%%%%%%%%%%%%%%%%%%%%%%%%%%%%%%%%%%%%%%%%%%%%%%%%%%%%%%%
\subsection{Calibration Analysis}
\label{sec:exp:calibration}

\begin{table}[h]
    \centering
    {\renewcommand{\arraystretch}{1.0}
    \begin{adjustbox}{width=.9\linewidth}
        \begin{tabular}{lcccc}
\toprule
\textbf{Model} & \textbf{ArXiv} & \textbf{Reddit} & \textbf{Yelp} & \textbf{News} \\
\midrule
LLP & 0.005 & 0.009 & 0.010 & 0.007 \\
CLP & 0.000 & 0.006 & 0.007 & 0.003 \\
RoBERTa & 0.002 & 0.006 & 0.033 & 0.002 \\
\bottomrule
\end{tabular}

    \end{adjustbox}}
    \caption{Expected Calibration Error (ECE) analysis comparing MGT probes against RoBERTa on DetectRL~\cite{wu2024detectrl}.}
    \label{tab:t_ece}
\end{table}

Table~\ref{tab:t_ece} presents a small-scale calibration analysis of the MGT probes, comparing their Expected Calibration Error (ECE) against RoBERTa~\cite{liu2019roberta}. The results show that both probes are well calibrated across DetectRL~\cite{wu2024detectrl} subsets and, in several settings, outperform a fully fine-tuned RoBERTa model in calibration.

We emphasize that this analysis is performed on the projection scores produced by the linear probes. These results provide additional evidence that the probe outputs are well calibrated and support their potential use for fine-grained AI-edit detection.

%%%%%%%%%%%%%%%%%%%%%%%%%%%%%%%%%%%%%%%%%%%%%%%%%%%%%%%%%%%%%%%%%%%%%%%%
% ABLATIONS
%%%%%%%%%%%%%%%%%%%%%%%%%%%%%%%%%%%%%%%%%%%%%%%%%%%%%%%%%%%%%%%%%%%%%%%%
\section{Ablation Studies}
\label{app:ablation}

Lastly, we conduct a series of ablation experiments for both probing variants LLP and CLP.

\subsection{LLP}

\subsubsection{Model Size and Architecture Ablations}

\begin{table}[h]
    \centering
    {\renewcommand{\arraystretch}{1.0}
    \begin{adjustbox}{width=.8\linewidth}
        \begin{tabular}{lcccc}
\toprule
 & \multicolumn{4}{c}{\textbf{TSM~\citep{quaremba2026tsm}}} \\
\cmidrule(lr){2-5}
\textbf{Setting} & \textbf{FP} & \textbf{PE} & \textbf{SUM} & \textbf{TST} \\
\midrule
Baseline & 0.971 & 0.942 & 0.978 & 0.896 \\
\addlinespace
\multicolumn{5}{l}{\textbf{Token Aggregation}} \\
\hspace*{1em}Pooling & -0.019 & -0.155 & -0.021 & -0.085 \\
\addlinespace
\multicolumn{5}{l}{\textbf{Layer Selection}} \\
\hspace*{1em}First layer & -0.388 & -0.316 & -0.431 & -0.355 \\
\hspace*{1em}Last layer & -0.036 & -0.077 & -0.038 & -0.065 \\
\addlinespace
\multicolumn{5}{l}{\textbf{Model}} \\
\hspace*{1em}Llama-3B & +0.001 & -0.012 & +0.006 & +0.000 \\
\hspace*{1em}Llama-1B & -0.011 & -0.034 & +0.004 & -0.011 \\
\hspace*{1em}Qwen-32B & +0.020 & +0.029 & +0.013 & +0.048 \\
\hspace*{1em}Qwen-8B & +0.016 & +0.016 & +0.010 & +0.031 \\
\hspace*{1em}Qwen-4B & +0.015 & +0.011 & +0.007 & +0.029 \\
\hspace*{1em}Qwen-0.6B & -0.012 & -0.036 & -0.005 & -0.022 \\
\addlinespace
\multicolumn{5}{l}{\textbf{Regularization Penalty}} \\
\hspace*{1em}C=0.01 & -0.001 & +0.001 & +0.000 & +0.001 \\
\hspace*{1em}C=0.1 & +0.000 & +0.001 & +0.000 & +0.002 \\
\hspace*{1em}C=10 & +0.001 & +0.001 & +0.000 & +0.002 \\
\addlinespace
\multicolumn{5}{l}{\textbf{PCA Activations}} \\
\hspace*{1em}k=10 & -0.064 & -0.117 & -0.046 & -0.120 \\
\hspace*{1em}k=50 & -0.009 & -0.015 & -0.008 & -0.022 \\
\hspace*{1em}k=150 & +0.004 & +0.005 & +0.001 & +0.011 \\
\hspace*{1em}k=200 & +0.003 & +0.006 & -0.001 & +0.011 \\
\hspace*{1em}k=250 & +0.003 & +0.006 & -0.002 & +0.010 \\
\hspace*{1em}No PCA & +0.016 & +0.010 & +0.007 & +0.014 \\
\bottomrule
\end{tabular}

    \end{adjustbox}}
    \caption{Ablation study of LLP across five design choices.}
    \label{tab:t_ablations}
\end{table}

Table~\ref{tab:t_ablations} reports AUC scores for the LLP variant defined in Section~\ref{sec:method}, covering five design choices: (\textit{1}) token aggregation (last token vs.\ mean pooling), (\textit{2}) layer selection (first layer, last layer, or layer-wise aggregation), (\textit{3}) model architecture and size, (\textit{4}) regularization strength, and (\textit{5}) the number of PCA components.
We restrict this analysis to TSM, as it consistently proved to be the most challenging benchmark throughout our experiments.

First, probing only the final token yields moderate but consistent improvements over mean pooling.
This likely reflects the fact that the final token aggregates information from the entire sequence through causal attention.

Second, using only the first or last layer instead of aggregating projection scores across layers consistently reduces performance.
This finding aligns with~\citet{nordby2026ensemble}, who show that ensembles of probes outperform single-layer probes.
We do not include a single middle-layer probe, as TextFluoroscopy~\cite{yu2024fluo} already optimizes for the best-performing layer, and our results demonstrate that LLP consistently outperforms this approach.

Third, we evaluate the impact of model architecture and size by replacing LLaMA with Qwen~\cite{yang2025qwen3}.
We find that Qwen-8B consistently improves performance over LLaMA-8B.
Notably, even Qwen-4B achieves slightly stronger results, despite its smaller size.
These findings suggest that architecture may play a more important role than parameter count and motivate future work on how model design influences the linear separability of MGT and HWT.
Within the LLaMA family, reducing model size leads to only modest performance degradation.

Fourth, varying the strength of the $L_2$ regularization penalty has virtually no effect on performance, suggesting that the learned MGT direction is stable and readily recoverable.

Finally, motivated by our finding that MGT and HWT are linearly separable in low-dimensional subspaces (\S\ref{sec:mgt_rep}), we vary the number of principal components used to project activations before training probes.
Increasing the dimensionality beyond $k=100$ yields only marginal and inconsistent improvements.
This result further supports our representation analysis, indicating that MGT signals are encoded in relatively low-dimensional regions of the activation space.

\subsubsection{Detailed PCA vs no-PCA comparison}

\begin{table}[H]
    \centering
    {\renewcommand{\arraystretch}{1.0}
    \begin{adjustbox}{width=1\linewidth}
        \begin{tabular}{lcccccccccccc}
\toprule
 & \multicolumn{4}{c}{Multisocial} & \multicolumn{4}{c}{DetectRL Domains} & \multicolumn{4}{c}{RAID Models} \\
\cmidrule(lr){2-5} \cmidrule(lr){6-9} \cmidrule(lr){10-13}
Mode & en & de & ru & zh & ArXiv & Reddit & Yelp & News & Cohere & GPT-4 & Llama & Mistral \\
\midrule
LLP (No PCA) & 0.963 & 0.933 & 0.915 & 0.950 & 1.000 & 1.000 & 1.000 & 1.000 & 0.969 & 0.992 & 0.995 & 0.989 \\
LLP (PCA) & 0.957 & 0.934 & 0.921 & 0.944 & 1.000 & 1.000 & 1.000 & 1.000 & 0.971 & 0.992 & 0.995 & 0.993 \\
\midrule
LLP (No PCA) $-$ LLP (PCA) & 0.006 & -0.001 & -0.006 & 0.006 & 0.000 & 0.000 & 0.000 & 0.000 & -0.002 & -0.001 & 0.001 & -0.004 \\
\bottomrule
\end{tabular}

    \end{adjustbox}}
    \caption{LLP ablation study comparing PCA and no-PCA variants. The PCA variant is the main specification, using 100 dimensions.}
    \label{tab:t_llp_pca_comparison}
\end{table}

A potential concern is that PCA may artificially facilitate linear separability, for example by removing noisy dimensions. While Table~\ref{tab:t_ablations} already shows nearly identical probe performance with and without PCA, we provide a more detailed comparison in Table~\ref{tab:t_llp_pca_comparison}. Specifically, we compare the no-PCA variant against the PCA variant using the top 100 principal components, as in our main experiments. The near-identical results indicate that PCA does not meaningfully affect linear separability. Rather, it primarily serves as a dimensionality-reduction step that reduces the computational and memory requirements of the probes.

\subsection{CLP}

\subsubsection{Model Size and Architecture Ablations}

\begin{table}[H]
    \centering
    {\renewcommand{\arraystretch}{1.0}
    \begin{adjustbox}{width=.8\linewidth}
        % \begin{tabular}{lcccc}
% \toprule
%  & \multicolumn{4}{c}{\textbf{TSM~\citep{quaremba2026tsm}}} \\
% \cmidrule(lr){2-5}
% \textbf{Setting} & \textbf{FP} & \textbf{PE} & \textbf{SUM} & \textbf{TST} \\
% \midrule
% Baseline & 0.973 & 0.920 & 0.970 & 0.890 \\
% \addlinespace
% \multicolumn{5}{l}{\textbf{Token Aggregation}} \\
% \hspace*{1em}Pooling & -0.008 & -0.088 & +0.001 & -0.036 \\
% \addlinespace
% \multicolumn{5}{l}{\textbf{Model}} \\
% \hspace*{1em}Llama-3B & -0.001 & -0.019 & -0.002 & -0.021 \\
% \hspace*{1em}Llama-1B & -0.016 & -0.033 & -0.002 & -0.015 \\
% \hspace*{1em}Qwen-8B & +0.012 & +0.001 & +0.010 & +0.020 \\
% \hspace*{1em}Qwen-4B & +0.005 & +0.018 & +0.010 & +0.009 \\
% \addlinespace
% \multicolumn{5}{l}{\textbf{Regularization Penalty}} \\
% \hspace*{1em}C=0.01 & +0.001 & +0.016 & +0.005 & +0.003 \\
% \hspace*{1em}C=0.1 & +0.001 & +0.009 & +0.003 & +0.000 \\
% \hspace*{1em}C=10 & -0.001 & -0.001 & -0.000 & -0.011 \\
% \addlinespace
% \multicolumn{5}{l}{\textbf{PCA Activations}} \\
% \hspace*{1em}k=10 & -0.036 & -0.010 & -0.009 & -0.056 \\
% \hspace*{1em}k=50 & -0.004 & +0.000 & -0.010 & -0.014 \\
% \hspace*{1em}k=150 & -0.002 & +0.003 & -0.000 & +0.002 \\
% \hspace*{1em}k=200 & +0.001 & +0.004 & +0.001 & +0.003 \\
% \hspace*{1em}k=250 & +0.003 & +0.004 & +0.002 & +0.003 \\
% \hspace*{1em}No PCA & +0.006 & +0.019 & +0.005 & +0.006 \\
% \bottomrule
% \end{tabular}

\begin{tabular}{lcccc}
\toprule
 & \multicolumn{4}{c}{\textbf{TSM~\citep{quaremba2026tsm}}} \\
\cmidrule(lr){2-5}
\textbf{Setting} & \textbf{FP} & \textbf{PE} & \textbf{SUM} & \textbf{TST} \\
\midrule
Baseline & 0.973 & 0.920 & 0.970 & 0.890 \\
\addlinespace
\multicolumn{5}{l}{\textbf{Token Aggregation}} \\
\hspace*{1em}Pooling & -0.008 & -0.088 & +0.001 & -0.036 \\
\addlinespace
\multicolumn{5}{l}{\textbf{Model}} \\
\hspace*{1em}Llama-3B & -0.001 & -0.019 & -0.002 & -0.021 \\
\hspace*{1em}Llama-1B & -0.016 & -0.033 & -0.002 & -0.015 \\
\hspace*{1em}Qwen-32B & +0.016 & +0.029 & +0.012 & +0.025 \\
\hspace*{1em}Qwen-8B & +0.012 & +0.001 & +0.010 & +0.020 \\
\hspace*{1em}Qwen-4B & +0.005 & +0.018 & +0.010 & +0.009 \\
\hspace*{1em}Qwen-0.6B & -0.012 & -0.016 & -0.003 & -0.043 \\
\addlinespace
\multicolumn{5}{l}{\textbf{Regularization Penalty}} \\
\hspace*{1em}C=0.01 & +0.001 & +0.016 & +0.005 & +0.003 \\
\hspace*{1em}C=0.1 & +0.001 & +0.009 & +0.003 & +0.000 \\
\hspace*{1em}C=10 & -0.001 & -0.001 & -0.000 & -0.011 \\
\addlinespace
\multicolumn{5}{l}{\textbf{PCA Activations}} \\
\hspace*{1em}k=10 & -0.036 & -0.010 & -0.009 & -0.056 \\
\hspace*{1em}k=50 & -0.004 & +0.000 & -0.010 & -0.014 \\
\hspace*{1em}k=150 & -0.002 & +0.003 & -0.000 & +0.002 \\
\hspace*{1em}k=200 & +0.001 & +0.004 & +0.001 & +0.003 \\
\hspace*{1em}k=250 & +0.003 & +0.004 & +0.002 & +0.003 \\
\hspace*{1em}No PCA & +0.015 & +0.031 & +0.012 & +0.026 \\
\bottomrule
\end{tabular}

    \end{adjustbox}}
    \caption{Ablation study of CLP across four design choices.}
    \label{tab:t_ablations_meta}
\end{table}

Table~\ref{tab:t_ablations_meta} presents CLP ablations across four design choices: (\textit{1}) token aggregation (last token vs.\ mean pooling), (\textit{2}) model architecture and size, (\textit{3}) regularization strength, and (\textit{4}) the number of PCA components.

Overall, we observe trends similar to those for LLP in Table~\ref{tab:t_ablations}. Using the last token consistently outperforms mean pooling. Smaller Llama variants reduce performance, whereas Qwen yields modest but consistent improvements. The regularization penalty has only a minor effect at $\lambda=0.001$ and a negligible impact for other values. Finally, as with LLP, 100 PCA components provide the best trade-off between performance and dimensionality, with no meaningful gains from using additional components.

\subsubsection{Detailed PCA vs no-PCA comparison}

\begin{table}[H]
    \centering
    {\renewcommand{\arraystretch}{1.0}
    \begin{adjustbox}{width=1\linewidth}
        \begin{tabular}{lcccccccccccc}
\toprule
 & \multicolumn{4}{c}{Multisocial} & \multicolumn{4}{c}{DetectRL Domains} & \multicolumn{4}{c}{RAID Models} \\
\cmidrule(lr){2-5} \cmidrule(lr){6-9} \cmidrule(lr){10-13}
Mode & en & de & ru & zh & ArXiv & Reddit & Yelp & News & Cohere & GPT-4 & Llama & Mistral \\
\midrule
CLP (PCA) & 0.951 & 0.920 & 0.906 & 0.951 & 1.000 & 1.000 & 1.000 & 1.000 & 0.965 & 0.991 & 0.995 & 0.987 \\
CLP (No PCA) & 0.962 & 0.937 & 0.919 & 0.951 & 1.000 & 1.000 & 1.000 & 1.000 & 0.967 & 0.989 & 0.995 & 0.987 \\
\midrule
CLP (PCA) $-$ CLP (No PCA) & -0.011 & -0.017 & -0.013 & 0.000 & 0.000 & 0.000 & -0.000 & 0.000 & -0.002 & 0.002 & -0.000 & 0.000 \\
\bottomrule
\end{tabular}

    \end{adjustbox}}
    \caption{CLP ablation study comparing PCA and no-PCA variants. The PCA variant is the main specification, using 100 dimensions.}
    \label{tab:t_clp_pca_comparison}
\end{table}

Table~\ref{tab:t_clp_pca_comparison} reports the same PCA versus no-PCA comparison for CLP. Consistent with the LLP results, performance is nearly identical across both variants, indicating that PCA does not meaningfully affect linear separability for either probe.

\end{document}